\documentclass[acmtog,authorversion,nonacm]{acmart}

\usepackage{multirow}
\newcommand{\refFig}[1]{Fig.~\ref{fig:#1}}
\newcommand{\refSec}[1]{Sec.~\ref{sec:#1}}
\newcommand{\refTbl}[1]{Tbl.~\ref{tbl:#1}}
\newcommand{\refEq}[1]{Eq.~\ref{eq:#1}}
\newcommand{\refSupp}[1]{Appendix~\ref{supp:#1}}

\newcommand{\mycomment}[1]{}

\definecolor{darkred}{rgb}{0.6,0,0}
\definecolor{darkblue}{rgb}{0.0,0,0.5}
\definecolor{green}{rgb}{0.0,0.5,0}
\definecolor{blue}{rgb}{0,0,0.75}
\definecolor{orange}{rgb}{1,0.6,0.2}
\definecolor{red}{rgb}{1,0,0}
\definecolor{purplish}{rgb}{0.6,0,0.7}

\def\dsmalours{\emph{SMAL\textsuperscript{eval}}}
\def\damass{\emph{SMPL\textsuperscript{eval}}}

\def\dsmal{\emph{SMAL}}
\def\dsurreal{\emph{SURREAL}}

\def\dshrechumans{\emph{SHREC'19}}
\def\dshrecanimals{\emph{SHREC'20}}
\def\dtosca{\emph{TOSCA}}

\def\reals{\mathbb{R}}

\def\mesh{\mathcal{M}}

\usepackage{subcaption}

\begin{document}

\title{Surface-Aware Distilled 3D Semantic Features}

\author{Lukas Uzolas}
\orcid{0000-0002-6857-8795}
\affiliation{\institution{Delft University of Technology}
\city{Delft}
\country{Netherlands}}
\email{l.uzolas@tudelft.nl}

\author{Elmar Eisemann}
\orcid{0000-0003-4153-065X}
\affiliation{\institution{Delft University of Technology}
\city{Delft}
\country{Netherlands}}
\email{e.eisemann@tudelft.nl}

\author{Petr Kellnhofer}
\orcid{0000-0001-7881-2285}
\affiliation{\institution{Delft University of Technology}
\city{Delft}
\country{Netherlands}}
\email{p.kellnhofer@tudelft.nl}

\renewcommand{\shortauthors}{Uzolas et al.}

\begin{abstract}
Many 3D tasks such as pose alignment, animation, motion transfer, and 3D reconstruction rely on establishing correspondences between 3D shapes. This challenge has recently been approached by pairwise matching of semantic features from pre-trained vision models. However, despite their power, these features struggle to differentiate instances of the same semantic class such as ``left hand'' versus ``right hand'' which leads to substantial mapping errors. To solve this, we learn a surface-aware embedding space that is robust to these ambiguities while facilitating shared mapping for an entire family of 3D shapes. Importantly, our approach is self-supervised and requires only a small number of unpaired training meshes to infer features for new possibly imperfect 3D shapes at test time. We achieve this by introducing a contrastive loss that preserves the semantic content of the features distilled from foundational models while disambiguating features located far apart on the shape's surface. 
We observe superior performance in correspondence matching benchmarks and enable downstream applications including 2D-to-3D and 3D-to-3D texture transfer, in-part segmentation, pose alignment, and motion transfer in low-data regimes.
Unlike previous pairwise approaches, our solution constructs a joint embedding space, where both seen and unseen 3D shapes are implicitly aligned without further optimization. The code is available at \url{https://graphics.tudelft.nl/SurfaceAware3DFeatures}.
\end{abstract}  



\begin{teaserfigure}
  \includegraphics[width=\textwidth]{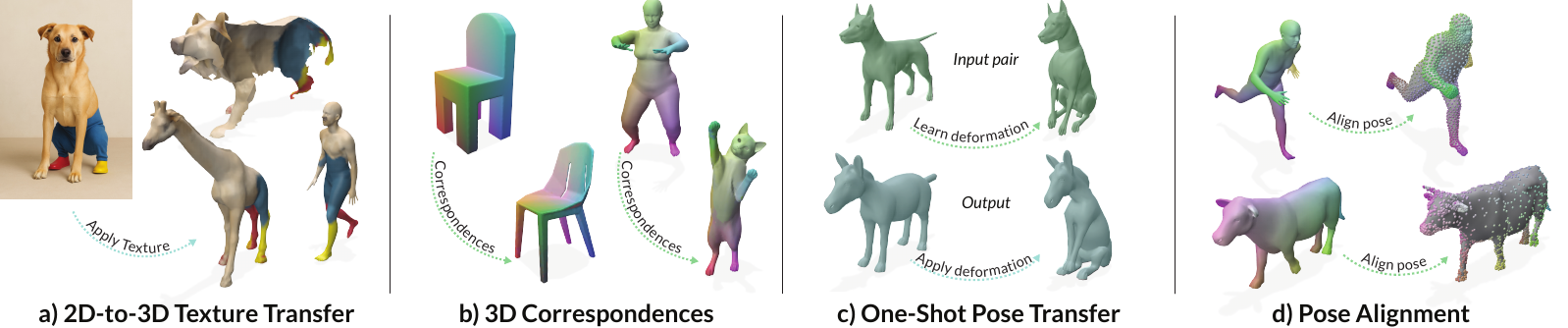}
  \caption{We introduce a \emph{surface-aware} feature embedding space based on 2D pre-trained foundational models. In contrast to related works, our embedding space separates instances of the same semantic class (e.g. right vs. left instances for ``hand/paw'') which facilitates many downstream applications: a) Texturing of even incomplete 3D shapes based on a 2D image (the input image was produced by ChatGPT), b) 3D correspondences between non-isometric shapes, c) Re-posing of meshes based on a single source and target pair, d) Pose alignment of 3D meshes with dense and sparse point correspondences.
  }
  \Description{Teaser.}
  \label{fig:teaser}
\end{teaserfigure}

\maketitle


\section{Introduction}
\label{sec:intro}
Mapping 3D shapes into a shared space guaranteeing mutual correspondences is important for many applications, including 3D registration, pose alignment, motion transfer, as well as static and dynamic 3D reconstruction.
Historically, geometric descriptors have been used to determine matches between pairs of 3D shapes under isometric deformations, but they struggle with non-isometric deformations~\cite{sun2009concise,tombari2010unique,aubry2011wave}.
In contrast, neural features, stemming from pre-trained 2D vision models, have recently achieved great success in identifying correspondences between pairs of vastly different shapes~\cite{tang2023emergent,luo2023diffusion,zhang2024tale,wimmer2024back}, such as mapping from cats to lions.
In this paper, we make another important step by moving from one-to-one pairwise shape correspondence matching to a joint embedding space establishing many-to-many shape correspondences.

Despite their inter-class robustness, neural features often struggle to disambiguate between instances of the same class like ``left hand'' and ``right hand'' (see ~\refFig{pca}).
Such mismatches can lead to substantial errors in downstream applications (see \refSec{applications}).
Recent research has demonstrated that these features contain global pose information and that disambiguation is possible in a 2D scenario~\cite{zhang2024telling}.
However, achieving the same effect on distilled 3D features is not trivial, especially in a low-data regime, which is prevalent in 3D, where data acquisition and labeling is difficult.

Our work improves 3D neural features distilled from pre-trained 2D vision models by embedding them into a space disambiguating intraclass instances. 
We achieve this without large annotated datasets using a self-supervised learning scheme guided by in-shape geodesic distances without the need for shape pairs.
Training with a limited number of 3D meshes, our method produces space of \emph{surface-aware features} establishing multi-faceted correspondences for diverse new shapes without any further fine-tuning.
In quantitative and qualitative comparisons to prior work, we demonstrate superior suitability of these features to serve as robust descriptors for matching and as building blocks for solving other tasks.
Since geodesic distances are not used during inference, our method has only a minimal overhead from its shallow neural encoder and its point-wise nature makes it robust to varying mesh complexity or shape incompleteness.
Finally, the encoder preserves compatibility with per-pixel image features, and hence, also naturally establishes robust 2D-to-3D mappings.

In summary, we make the following contributions: 
1. We introduce a novel contrastive loss for self-supervised distillation of 3D features. 
2. We quantitatively demonstrate the effectiveness of our surface-aware features in pose transfer, correspondence matching and skinning weight regression.
3. We showcase versatility of our approach in additional downstream applications including pose alignment, instance-based part segmentation and 2D-to-3D or 3D-to-3D texture transfer.
4. We show the versatility of our features when matching many-to-many shapes of not only humanoids and animals but also other classes.
\section{Related Works}
\label{sec:related_works}
Our method utilizes contrastive learning to embed semantic features from foundational models to a space enabling robust n-to-n 3D shape matching. In this section, we discuss prior work in these three areas.

\subsection{Image-based features for 3D shapes}
Image-based features emerge in large visual models for 2D image tasks.
Self-supervised features from Vision Transformers, such as DINO-ViT~\cite{caron2021emerging} and DINOv2~\cite{oquab2024dinov}, locally encode semantic information useful for segmentation~\cite{caron2021emerging} or image-to-image correspondence matching~\cite{amir2021deep}.
SD-DINO~\cite{zhang2023tale} adds complementary features from the diffusion-based image synthesis model Stable Diffusion~\cite{rombach2022high}.
Lifting these features to 3D has enabled the self-supervised construction of canonical surface maps~\cite{shtedritski2024shic}, transfer of appearance between 3D shapes~\cite{fischer2024nerf}, 3D animation~\cite{uzolas2024motiondreamer}, keypoint detection~\cite{wimmer2024back} or matching of surface correspondences~\cite{dutt2024diffusion,morreale2024neural,chen2025dvmatcherdeformationbasednonrigidpoint}.
However, despite their semantic versatility, disambiguating between intraclass instances, such as left and right hands, remains challenging but possible, as shown in a recent 2D image study~\cite{zhang2024telling}. This motivates our 3D shape descriptors for resolving in-
stance ambiguity.

Prior work tackled this ambiguity by mapping shapes to a spherical template~\cite{mariotti2024improving}, which is difficult for complex shapes including humans. Alternatively, Liu~et~al.~\shortcite{liu2025stable} recovered non-isometric correspondences from a 2D semantic flow learned from vision features.
Instead, we adapt Diff3F features~\cite{dutt2024diffusion} in 3D space and resolve the ambiguity through contrastive learning enforcing geodesic distances.

Geodesic distances have previously supported point cloud analysis~\cite{he2019geonet} and more recently, NIE~\cite{jiang2023neural} and the concurrent work DV-Matcher~\cite{chen2025dvmatcherdeformationbasednonrigidpoint} similarly utilize geodesic distances for feature disambiguation.
Yet, the previously mentioned methods rely on aligned mesh pairs or a learned alignment, while our method learns purely from intrinsic properties of individual shapes.
This eases adaptation to less common classes beyond humanoids and animals (\refFig{tosca_shapenet_examples}) and cross-class mappings (\refFig{transfer_centroids}).

Beyond vision-only models, multimodal large language models have recently been effective in image and 3D shape analysis including keypoint labeling \cite{gong2024zerokey} and shape co-segmentation \cite{abdelreheem2023zero}. 
In our work, we focus on vision-only models because of their simplicity, but we consider a model combination a promising research direction.

\subsection{Contrastive Learning}
Contrastive learning embeds similar samples close to each other while keeping dissimilar samples apart.
This can be achieved directly by minimizing and maximizing embedding distances for positive and negative pair samples, respectively~\cite{chopra2005learning, hadsell2006dimensionality, weinberger2009distance, schroff2015facenet} or indirectly, such as by optimizing performance in an auto-regressive task~\cite{oord2018representation}.
Training pairs can be obtained by data augmentation~\cite{chen2020simple}, from memory banks~\cite{wu2018unsupervised,he2020momentum}, or by clustering~\cite{caron2018deep,caron2020unsupervised}.
Learning with cross-domain labels yields joint embeddings, as demonstrated by CLIP~\cite{radford2021learning} for text and images.
Contrastive learning was applied to learn end-to-end pose transfer from multiple unregistered meshes of the same identity in different poses~\cite{sun2023mapconnet}.
We design our contrastive loss to disambiguate intra-class instances guided by a geodesic metric, while learning from intrinsic properties of individual meshes rather than same-identity shape pairs.

\subsection{Shape correspondences}
\label{sec:rel_correspondences}
\paragraph{Point-to-Point}
Classical shape registration methods directly minimize global~\cite{besl1992method} or local~\cite{brown2007global} inter-shape distances making them susceptible to local minima~\cite{yang2015go}.
This motivates the design of more informative local geometric descriptors~\cite{sun2009concise,tombari2010unique,aubry2011wave}.
These can alternatively be learned~\cite{corman2014supervised,guo20153d} from voxelized patches~\cite{zeng20173dmatch,gojcic2019perfect,attaiki2023generalizable} or from point clouds~\cite{elbaz20173d,yew20183dfeat,deng2018ppfnet,deng2018ppffoldnet,deng2023se}.
The learning can be supervised by labels~\cite{corman2014supervised} or achieved without them~\cite{elbaz20173d,groueix20183d,lang2021dpc,zeng2021corrnet3d}.
Our method falls into the latter category, as our contrastive loss motivates our encoder to separate instances by approximating geodesic distances~\cite{xia2021geodesicembedding} without training data labels.
This is conceptually similar to previous methods for near-isometric shape deformations~\cite{memoli2005theoretical,halimi2019unsupervised,shamai2017geodesic}. 
However, we distinctly do not measure geodesic distortions between shape pairs, and therefore we do not limit our method to isometric deformations, and we do not compute any geodesics during inference.
Instead, we only use the geodesics to disambiguate information already available in the image-based features, which is critical for our results.

The correspondences can be recovered from descriptors by a matching~\cite{fischler1981random}, directly regressed~\cite{lu2019deepvcp,wang2019deep} or established on parametric templates~\cite{groueix20183d,deprelle2019learning}.
Here, we focus on the descriptors themselves, and we show several different application scenarios in \protect\refSec{applications}.

\paragraph{Surface mapping}
Functional Maps (FMs)~\cite{ovsjanikov2012functional} allow for matching on a surface.
FMs are real-valued surface functions in the space of Laplace-Bertrami eigenfunctions, supporting linear transformations between shapes.
Constrained to match surface descriptors for each shape~\cite{sun2009concise,tombari2010unique,aubry2011wave} they allow extracting point-wise correspondences~\cite{ovsjanikov2012functional,rodola2015point}.
These functions can also be learned~\cite{litany2017deep} often with little or no supervision \cite{roufosse2019unsupervised,donati2020deep,ginzburg2020cyclic,halimi2019unsupervised,sun2023spatially}. 
Extrinsic alignment can support nonisotropic deformations~\cite{eisenberger2020smooth,eisenberger2020deep}.
In this work, we focus on improving features for direct point-to-point matching in the spatial domain, but we later demonstrate a combination of our features with FM.

\begin{figure*}[ht]
    \centering
    \includegraphics[width=1\textwidth]{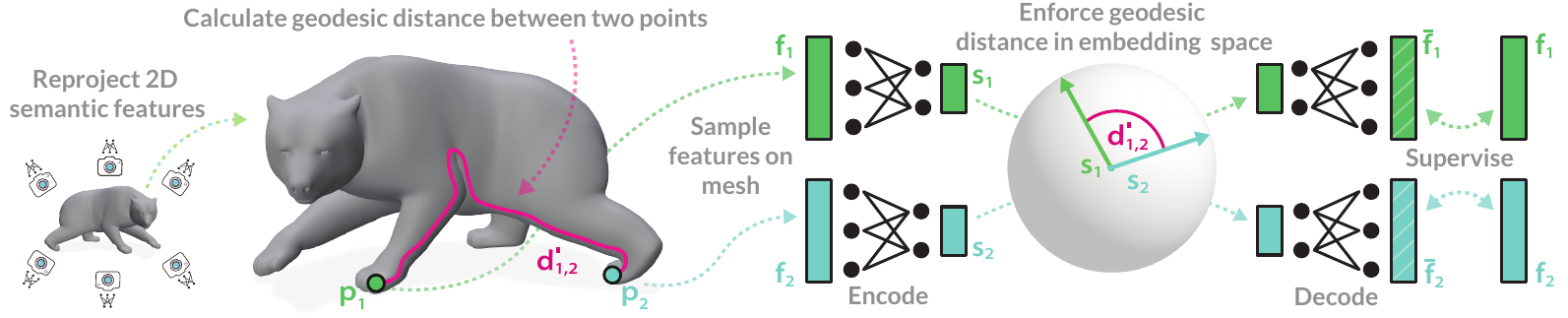}
    \caption{
    Overview of our method.
    We feed images of a 3D shape rendered from multiple viewpoints to a pre-trained 2D vision model and extract features that are then projected back onto surface points $\mathbf{p}_i$ and aggregated into per-point features $\mathbf{f}_i$ (\refSec{preliminaries}).
    Next, we pointwise embed the \emph{base features} $\mathbf{f}_i$ into our \emph{surface-aware features} $\mathbf{s}_i$ residing in a lower-dimensional space learned using our contrastive loss preserving geodesic distances $d_{i,j}$ and a reconstruction loss matching decoded features $\mathbf{\bar{f}}_i$ to $\mathbf{f}_i$ (\refSec{method}). 
    The \emph{surface-aware features} $\mathbf{s}_i$ serve as robust descriptors for correspondence matching (\refSec{experiments}) and base blocks for many down-stream applications (\refSec{applications}).
    }
    \label{fig:method_overview}
\end{figure*}

\section{Preliminaries}
\label{sec:preliminaries}
We build upon methods that aggregate features from pre-trained 2D vision models on 3D meshes~\cite{dutt2024diffusion,wimmer2024back,morreale2024neural,chen2025dvmatcherdeformationbasednonrigidpoint}. In this section, we give a brief overview on these methods.

\subsection{Reprojection of 2D Features}
We represent a 3D shape as a triangular mesh with a tuple of $N$ vertices and $M$ triangular faces, that is, $\mesh{} := (\{\mathbf{p}_n \in \reals^3 | n = 1,...,N\}, \{\mathbf{t}_m \in \mathbb{N}^3| m = 1,...,M\})$.
The rendering function $R_{rgb}:(\mathcal{M}, \mathcal{C}) \rightarrow \mathbf{I}_{rgb}$ projects $\mesh{}$ into a camera $\mathcal{C}$ and outputs an image $\mathbf{I}_{rgb} \in \reals^{H\times W \times 3}$, with height $H$ and width $W$.
Optionally, texturing is possible in $R_{rgb}(.)$ or as a ControlNet~\cite{zhang2023adding} post-processing.
The image is then passed to a pre-trained vision model~\cite{caron2021emerging,oquab2024dinov,rombach2022high,zhang2023tale} to obtain dense semantic feature maps $\mathbf{F}\in \reals^{h\times w \times f}$ with $h, w, f$ as two spatial and one feature dimension.
Finally, per-vertex features $\mathbf{f}_n \in \reals^f$ are obtained by projective texture mapping of $\mathbf{F}$ onto $\mathcal{M}$.
To cover the whole surface, features are aggregated across multiple cameras, resulting in a set of features $\mathcal{F}_{\mathcal{M}}:=\{\mathbf{f}_n \in \reals^f | n = 1,...,N\}$.
Throughout this work, we refer to $\mathcal{F}_{\mathcal{M}}$ as the \emph{base features} on which our method is built.
The exact choice of $\mathcal{F}_{\mathcal{M}}$ is orthogonal to our contribution but must encode semantic information. 
To this extent, we use Diff3F~\cite{dutt2024diffusion} in this work.

\paragraph{Correspondence Matching}
Features $\mathcal{F}_{\mathcal{M}}$ have been shown to encode strong semantic information useful for correspondence matching~\cite{tang2023emergent,dutt2024diffusion}.
In the simplest case, the feature $\mathbf{f}_n\in \mathcal{F}_{\mathcal{T}}$ of a target mesh $\mathcal{T}$ that best matches the feature $\mathbf{f}_m \in \mathcal{F}_{\mathcal{S}}$ of a source mesh $\mathcal{S}$
is determined by maximizing the cosine similarity $\phi: \reals^f \times \reals^f \rightarrow \reals$:
\begin{equation}
    \phi(\mathbf{f}_i,\mathbf{f}_j) = \frac{\mathbf{f}_i^T\mathbf{f}_j}{\lVert \mathbf{f}_{i} \lVert_2 \lVert \mathbf{f}_{j} \lVert_2},
\end{equation}
\label{eq:cosine_similarity}
such that $\tau(\mathbf{p}_m) = \arg\max_{\mathbf{p}_n} \phi(\mathbf{p}_n\to\mathbf{f}_n,\mathbf{p}_m\to\mathbf{f}_m)$ is the best matching point.
However, the features $\mathcal{F}_{\mathcal{M}}$ do not differentiate between semantic instances well (see ~\refFig{qualitative_comparison}) which we address by learning robust \emph{surface-aware features} $\mathcal{S}_{\mathcal{M}}$.

\section{Method}
\label{sec:method}
Our goal is to learn an embedding resolving instance ambiguities of the \emph{base features} $\mathcal{F}_{\mathcal{M}}$ and obtain \emph{surface-aware features} $\mathcal{S}_{\mathcal{M}}$ (see \refFig{method_overview}).
We achieve this by training a point-based feature auto-encoder with a limited set of training meshes and our contrastive loss for self-supervision.
At test time, we can produce \emph{surface-aware features} for novel unseen shapes without additional fine-tuning.

\subsection{Setup}
\label{sec:method_setup}
Our method requires a potentially small set of training meshes $\mathbf{M}_t=\{\mesh{}_i | i = 1,...,K\}$, each associated with \emph{base features} $\mathcal{F}_{\mathcal{M}}$ obtained following \refSec{preliminaries} and normalized by a Euclidean norm such that $\forall \mathbf{f}_n \in \mathcal{F}_{\mathcal{M}}, \lVert\mathbf{f}_n\lVert_2 := 1$.

Unlike other approaches~\cite{jiang2023neural,deng2023se}, we do not require extrinsic canonical mesh alignment during training, because we rely only on intrinsic properties of individual meshes.
Similar considerations were previously made for Functional maps~\cite{ovsjanikov2012functional}.
Moreover, we inherit the rotation invariance of the \emph{base features} demonstrated by Dutt~et~al.~\shortcite[Supplement “Robustness to Rotation”]{dutt2024diffusion}, although we observe a performance degradation if meshes are upside down and we avoid this in our inputs (see \refSupp{diff3f_baselines}).

Furthermore, our encoder is point-based and does not rely on shape completeness or a consistent topology.
Both of these design choices favor generalization under transformations ranging from coordinate swap to shape reposing, and memory-unconstrained batch-based processing of even large shapes.

\subsection{Separating Front Paw from Back Paw}
\label{sec:method_separation}
Our embedding aims to separate multiple instances of the same class that are difficult to directly disambiguate in $\mathcal{F}_{\mathcal{M}}$.
For example, consider the two surface points, $\mathbf{p}_{1}$ and $\mathbf{p}_{2}$, on the bear's paws in \refFig{method_overview}.
The prevalent semantic significance of the ``paw'' concept hinders the separability of the corresponding \emph{base features} $\mathbf{f}_{1}$ and $\mathbf{f}_{2}$.
\refFig{pca} illustrates this for human arms and Diff3F features~\cite{dutt2024diffusion}.
To solve this, we train a point-wise feature autoencoder, producing our \emph{surface-aware features} $\mathcal{S}_{\mathcal{M}} \subset \reals^s$ in its embedding space.
We motivate the feasibility of separation by the prior observations that vision features additionally carry information about the global pose~\cite{zhang2024telling}.
We postulate that this enables our model to distinguish between part instances when guided by their intrinsic distance.
We adopt the geodesic distance $d_{1,2}$ between $\mathbf{p}_{1}$ and $\mathbf{p}_{2}$ for this purpose.
Following contrastive learning, we sample point pairs on a single shape to enforce $\phi(\mathbf{s}_{1}, \mathbf{s}_{2}) \approx d_{1,2}$ for $\mathbf{s}_{n} \in \mathcal{S}_{\mathcal{M}}$. 
We validate our choice of hyperspherical embedding space against Euclidean space in \refSec{ablations}.

\paragraph{Model}
\label{sec:method_training}
We train a \emph{base feature} encoder $\mathcal{E}(.)$, such that, following a normalization, $\mathbf{s}_n = \mathcal{E}(\mathbf{f}_n) / \lVert\mathcal{E}(\mathbf{f}_n)\lVert_2$ is a \emph{surface-aware feature} $\mathbf{s}_n \in \reals^s$ in a hypersphere embedding.
During training, we randomly sample an unpaired training mesh $\mathcal{M}\in\mathbf{M}_t$ with \emph{base features} $\mathcal{F}_{\mathcal{M}}$, which we encode pointwise to obtain $\mathcal{S}_{\mathcal{M}}$.

In each training iteration, we use furthest-point sampling to choose a random subset of $A$ anchor points $\mathbf{p}_a$ among the mesh vertices $\mathbf{p}_i \in \mesh{}$  
and compute geodesic distances $d_{n,a}$ for each pair of a mesh and anchor point. 
We additionally rescale $d_{n,a}$ to a maximum of one, such that $d'_{n,a} := d_{n,a} / \max_{n,a}(d_{n,a})$, which removes the dependency on the scale of the mesh.
We find this robust and leading to features later generalizing across morphologically equivalent shapes with different proportions (see an elephant vs. a giraffe in
\refFig{tosca_shapenet_examples}).

Subsequently, our contrastive loss preserves the rescaled geodesic distances in the embedding space:
\begin{equation}
    \mathcal{L}_{c} = \frac{1}{NA}\sum_{n}^N\sum_a^A \left| d'_{n,a} - \left(\frac{1 -\phi(\mathbf{s}_n, \mathbf{s}_a) }{2}\right)\right|.
\end{equation}
\label{eq:contrastive_loss}
This loss operates in a hyperspherical embedding space and utilizes cosine similarity mapped to the $[0,1]$ range.
Hereby, $\mathcal{L}_{c}$ penalizes features close in the embedding space but distant on the shape surface and vice versa.

Furthermore, we found it beneficial for the preservation of semantic information to train a feature decoder $\mathbf{\bar{f}}_n = \mathcal{D}(\mathbf{s}_n) / \lVert\mathcal{D}(\mathbf{s}_n)\lVert_2$ in an autoencoder fashion.
To this extent, we utilize a reconstruction loss:
\begin{equation}
    \mathcal{L}_{r} = \frac{1}{N} \sum_n  1 - \phi(\mathbf{f}_n, \mathbf{\bar{f}}_n).
\end{equation}
\label{eq:reconstruction_loss}
We train both the encoder and the decoder end-to-end with the combined loss $\mathcal{L} = w_r\mathcal{L}_{r}+w_c\mathcal{L}_{c}$, where a choice $w_r = w_c = 1$ works well in our tests. 
We do not observe an increase in performance with a higher $w_c$. 

Note that our training procedure, in contrast to related works~\cite{lang2021dpc,deng2023se,chen2025dvmatcherdeformationbasednonrigidpoint}, does not require target and source shape pairs.

\subsection{Implementation}
\label{sec:method_implementation}
During preprocessing, we rasterize our triangular meshes and precompute \emph{base features} for all vertices following \refSec{preliminaries}.
We implement our autoencoder in PyTorch2~\cite{ansel2024pytorch}
and use the Polyscope renderer~\cite{polyscope} for visualizations.
The encoder $\mathcal{E}$ is a Multilayer Perceptron (MLP) consisting of three blocks, where each block has two linear layers, SiLU activation~\cite{elfwing2018sigmoid}, and layer normalization~\cite{ba2016layer}. The first layer in each block employs a skip connection~\cite{he2016deep}, while the second reduces the dimensionality by a factor of two.
With Diff3F~\cite{dutt2024diffusion} as \emph{base features}, $\mathcal{E}$ reduces feature dimensionality from $f=2048$ to $s=256$. 
The decoder $\mathcal{D}$ is a mirrored copy of the encoder. We train our model on NVIDIA RTX 3090 for 50k iterations with the AdamW optimizer~\cite{loshchilov2017decoupled} and a learning rate of $0.0001$ which takes $\approx2$ hours.

We choose an exponential moving average~\cite{polyak1992acceleration} of the model with the lowest validation loss, without the need for any correspondence labels.
Geodesic distances for training are calculated on the fly with the heat method~\cite{crane2017heat} implemented in Geometry Central~\cite{geometrycentral}.
No geodesics are required during inference and the computational cost is determined by the Diff3F baseline with a only a negligible overhead from our shallow encoder $\mathcal{E}$.
For a shape with 10k vertices, this is less than 5 milliseconds on top of $\approx 4$ minutes from Diff3F.
Moreover, downstream tasks cost benefits from the smaller feature dimensionality.
Functional maps are calculated with the base algorithm~\cite{ovsjanikov2012functional}, provided by the Diff3F implementation~\cite{dutt2024diffusion}.

\section{Experiments}
\label{sec:experiments}
Here, we first motivate the benefits of our \emph{surface-aware feature} embedding space by visualizing its distribution.
Next, we evaluate their effectiveness in tasks with quantitative benchmarks including pose transfer, skinning weight regression and 3D correspondence matching. 
Finally, we analyze the impact of our design choices in an ablation experiment.

\paragraph{Training}
We train a single autoencoder on a joint dataset consisting of 49 animal samples from the \dsmal{} dataset~\cite{zuffi20173d} and 49 humans from the \dsurreal{} dataset~\cite{groueix20183d}. We choose 2 samples from each dataset for validation. We use this single shared model without any additional optimization for all experiments, unless stated otherwise.

\subsection{Exploration of Embedding Space}
To illustrate the effect of our contrastive loss on feature separation, we compare the 2D projections of the Diff3F~\cite{dutt2024diffusion} \emph{base features} with our \emph{surface-aware features}.

\begin{figure}[]
    \centering
    \includegraphics[width=\linewidth]{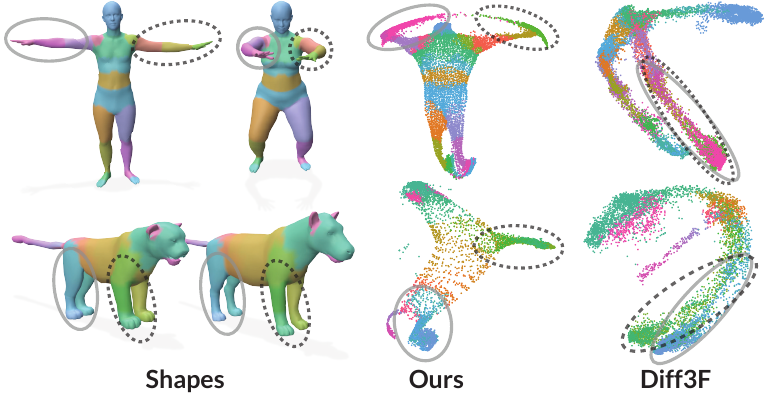}
    \caption{Two shapes (left) and a PCA-based 2D projections of their aggregated Diff3F \emph{base features} and our \emph{surface-aware features} (right). Notice the separation of limbs in our result compared to Diff3F.
    Our features originate from the same encoder for both shapes.
    The animal legs appear merged along the sagittal plane due limitations of the PCA projection, but they remain disambiguated in our feature space as demonstrated in \refFig{transfer_centroids}.
    }
    \label{fig:pca}
\end{figure}

\paragraph{Setup} 
We create two dataset \damass{} and \dsmalours{} unseen during training. 
The former consists of 50 randomly-sampled \textit{SMPL} \cite{loper2023smpl} shapes and poses from \textit{AMASS} \cite{mahmood2019amass}, while the latter consists of 50 randomly-sampled \dsmal{}~\cite{zuffi20173d,biggs2018creatures} shapes in canonical poses. 
For each sample, we obtain the \emph{base features} and \emph{surface-aware features} as described in \refSec{preliminaries} and \refSec{method}

\paragraph{Embedding} We project Diff3F features aggregated from \damass{} to two dimensions using principal component analysis (PCA). In \refFig{pca}, we visualize the projection for two selected shapes from the same dataset.
We repeat this with our \emph{surface-aware features}.
To avoid bias, we derive the visualized colors from the true SMPL \cite{loper2023smpl} skinning weights $\mathbf{w}_n\in \reals^{B}$ for both methods, where $B$ is the skinning weight dimension.
We repeat this process for \dsmalours.
In \refFig{pca}, our method yields an interpretable embedding that separates the leg and hand instances for animals and humans despite not having access to extrinsic $(x,y,z)$ point positions.
This validates the suitability of our features for downstream tasks and highlights the limitations of the Diff3F \emph{base features}.

\subsection{One-shot Pose Transfer}

\begin{figure}
    \centering
    \includegraphics[width=1\linewidth]{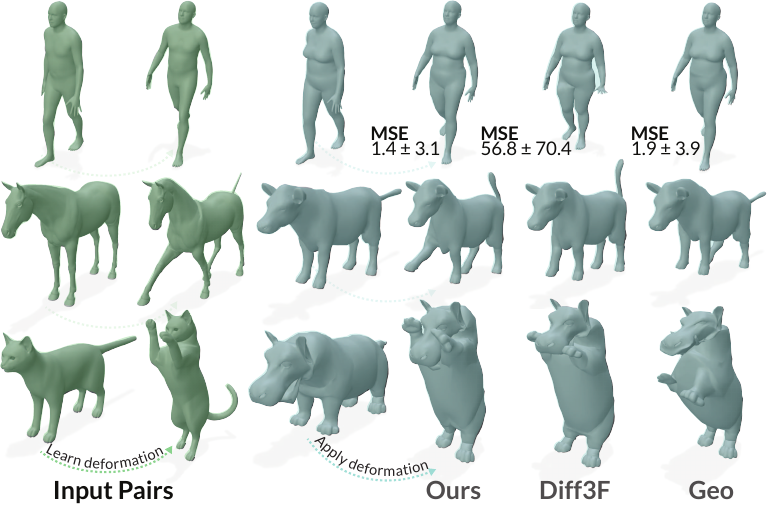}
    \caption{One-shot pose transfer using our features, Diff3F features, or Geometric descriptors. MSE $\times10^{-4}$ is reported for human shapes.}
    \label{fig:pose_transfer}
\end{figure}

We evaluate the performance of the \emph{surface-aware features} in a one-shot re-posing task for arbitrary 2-manifold meshes. We use our features to fit a Neural Jacobian Field (NJF)~\cite{aigerman2022neural} between poses of two input shapes and then apply it to re-pose a new target shape as described in \refSupp{pose_transfer}.

We sample 5 input shape pairs from \damass{} by choosing one challenging pose as a target and one random shape as an initial pose.
The remaining shape samples serve as test inputs for pose transfer with known ground truth.
We report MSE for 240 such test pairs (see upper row, \refFig{pose_transfer}).
Next, we repeat the same procedure with the \emph{base features} and with the Geometric descriptors  (\textsc{Geo}) \cite{aigerman2022neural}, consisting of the face centroid, face normals, and a Wave-Kernel Signature \cite{aubry2011wave}.
Finally, we provide additional qualitative results for transfer of animal poses from \dtosca{} to \dsmalours{} in \refFig{pose_transfer}.

We observe that our \emph{surface-aware features} outperform the \emph{base features} both quantitatively and qualitatively by correctly distinguishing individual posed limbs.
The Geometric descriptors also perform well for humans, but they struggle with larger input and output shape differences in animals.
Our features perform well in both cases.

\begin{figure}[]
    \centering
    \begin{tabular}{c c}
        \includegraphics[width=0.3\linewidth]{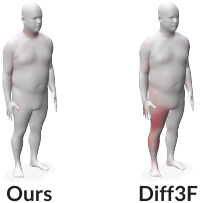} & 
        \raisebox{12mm}{
        \small
        \begin{tabular}{ccc}
            \hline
            & \damass & \dsmalours  \\
            \hline
            Ours & \textbf{.0014\footnotesize{±.0006}} & \textbf{.0048\footnotesize{±.0008}} \\
            Diff3F & .0058\footnotesize{±.0024} &  .0086\footnotesize{±.0028}\\
            Geo & .0210\footnotesize{±.0031} &  .0287\footnotesize{±.0017}\\
            \hline
        \end{tabular}
        }
    \end{tabular}
    \captionof{figure}{Mean Squared Error of skinning weight regression ($\downarrow$ is better) and its distribution across the SMPL mesh surface. }
    \label{fig:skinning_weight_regression}
\end{figure}

\subsection{Skinning Weight Regression}
We train a simple regressor to predict skinning weights of a kinematic model, based on a single training sample.

Our pointwise regressor $\mathcal{W}_\mathbf{s}(\mathbf{s}_n)$ consists of a linear layer and a Softmax activation and regresses skinning weights $\bar{\mathbf{w}}_n$ from our \emph{surface-aware features} or, in case of $\mathcal{W}_\mathbf{f}(\mathbf{f}_n)$ and $\mathcal{W}_\mathbf{g}(\mathbf{g}_n)$, from the \emph{base features} or the Geometric descriptors, respectively.
We train all models fivefold supervised with the Mean Squared Error (MSE) and true weights separately on the \damass{} and \    dsmalours{} datasets, and we report test MSE for the remaining unseen samples in the source datasets (see \refFig{skinning_weight_regression}).
Our features achieve lower errors and exhibit better robustness to instance ambiguities than the two alternatives. 
It is worth noting that the relatively lower dimensionality of the Geometric descriptors and \emph{base features} affects the number of regressor parameters, impact of which is not studied in this experiment. However, our conclusions also hold for a 2-layer MLP regressor with an equal hidden dimensionality of 106 which matches the dimensionality of the Geometric descriptors.

\subsection{Point-to-Point Correspondence Matching}
\label{sec:exp_correspondences}
Our features can be easily integrated into correspondence matching pipelines.
Therefore, we replicate the evaluation setup of Diff3F~\cite{dutt2024diffusion} and assess our \emph{surface-aware features} in a correspondence matching task on human and animal shapes.

\paragraph{Data}
For testing we use re-meshed versions of humans from \dshrechumans~\cite{melzi2019shrec,donati2020deep} and animals from both \dshrecanimals~\cite{dyke2020shrec} and the animal-only subset of \dtosca~\cite{bronstein2008numerical}.

\paragraph{Baselines} We compare our method against the unsupervised image-based Diff3F method~\cite{dutt2024diffusion}, which also provides our \emph{base features}, and against 3DCODED~\cite{groueix20183d}, DPC~\cite{lang2021dpc} and SE-OrNet~\cite{deng2023se}, which have been trained on thousands of samples, while our method is trained on less than 100 samples.

\paragraph{Metrics}
We report commonly used point correspondence metrics for $1024$ points per mesh~\cite{groueix20183d,lang2021dpc,deng2023se,dutt2024diffusion}\footnote{
We use the provided code and validate that we follow the same experimental procedure and metric definitions.
}.
The correspondence error measures a distance between the computed correspondence point $\tau(\mathbf{p}_n)$ (see \refSec{preliminaries}) and the ground-truth correspondence point $\mathbf{t}_n^{gt}$
: $err = \frac{1}{n} \sum_{\mathbf{p}_n\in \mathcal{S}} \lVert \tau(\mathbf{p}_n) - \mathbf{t}_n^{gt}\rVert_2^2$.
The accuracy is the fraction of points with an error below a threshold $\epsilon \in[0,1]$:  $acc(\epsilon) = \frac{1}{n} \sum_{\mathbf{p}_n\in \mathcal{S}} \mathbb{I}(\lVert \tau(\mathbf{p}_n) - t_n^{gt}\rVert_2 < \epsilon g)$, where $g$ is the maximal Euclidean distance in the target shape and $\mathbb{I}(.)$ is the indicator function.

\begin{figure*}[p]
    \centering
    \includegraphics[width=0.98\textwidth]{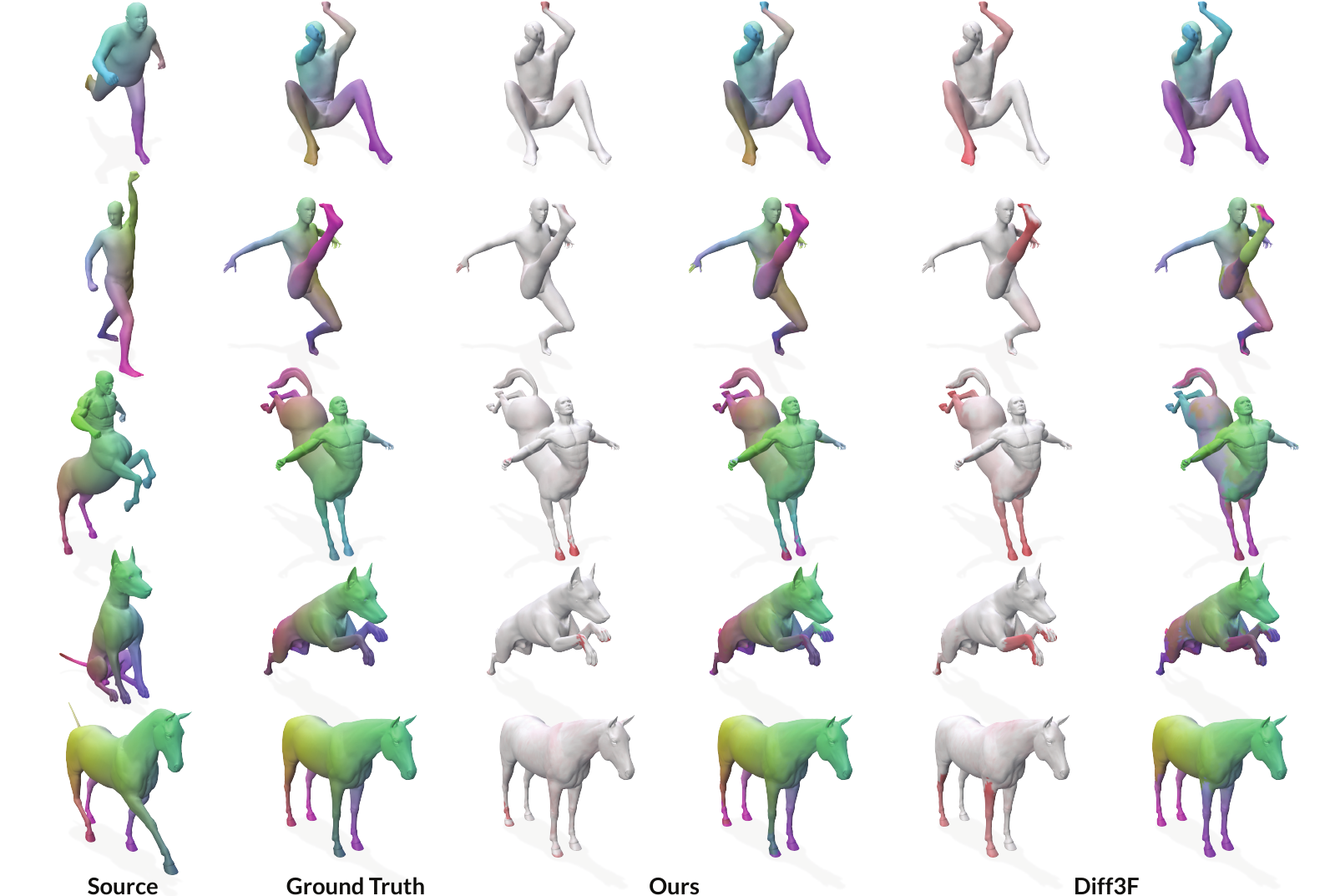}
    \caption{Qualitative comparison on the \dshrechumans{} and \dtosca{} datasets with dense true correspondence labels provided by their authors. 
    We show the source and target meshes with their ground truth correspondence labels (the two left-most columns) in comparison to correspondences computed using our \emph{surface-aware features} (the forth column) and Diff3F \emph{base features} (the right-most column). We further highlight the correspondence error on the mesh surface (the third and the fifth column). 
    The error colormap is normalized per sample by the maximal error over both methods to keep the error scale comparable across columns but not across rows.
    Our \emph{surface-aware features} notably improve separation of the limb instances.}
    \label{fig:qualitative_comparison}

    \begin{minipage}[t]{0.48\textwidth}
       \centering
        \includegraphics[width=\linewidth]{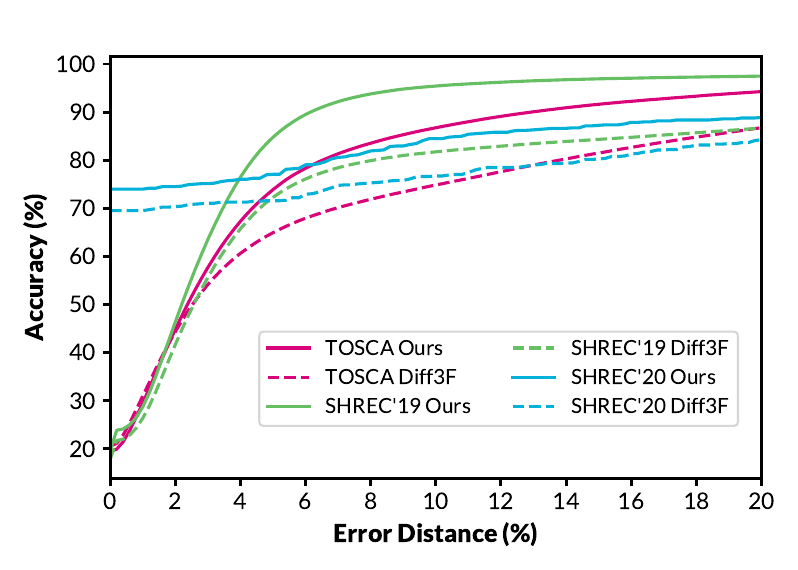}
        \caption{Correspondence accuracy ($\uparrow$ is better) at different error distance thresholds for our own and Diff3F features.}
        \label{fig:accuracy_graph}
      \end{minipage}\hfill%
      \begin{minipage}[t]{0.48\textwidth}
        \centering
        \includegraphics[width=\linewidth]{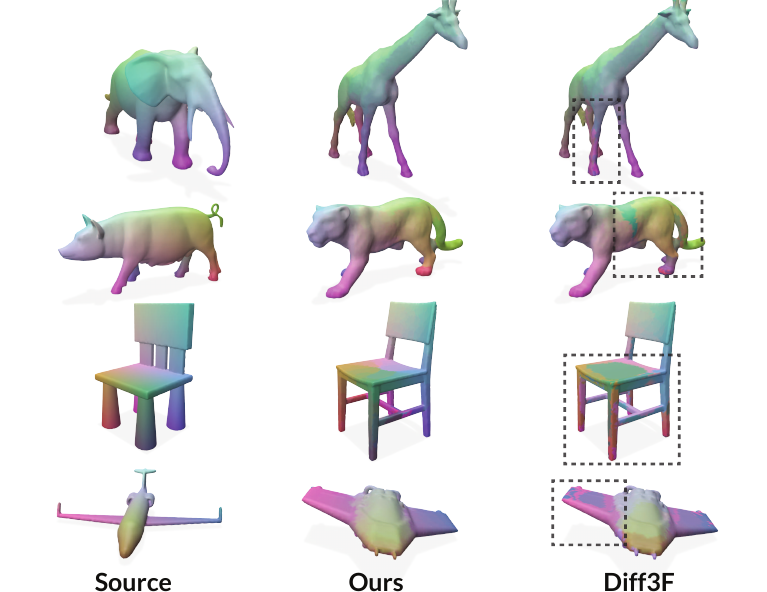}
        \caption{Qualitative comparison of correspondence matching on \dtosca{} and ShapeNet ~\cite{chang2015shapenet} (dense ground truth labels not available). 
        Source shape (left) matched to target (right) using our and Diff3F features.}
        \label{fig:tosca_shapenet_examples}
      \end{minipage}
        
\end{figure*}

\paragraph{Results}
We provide quantitative results in \refTbl{results_diff3f} and qualitative comparisons in \refFig{qualitative_comparison}. 
We find that our method achieves the lowest error on \dshrechumans{} and \dtosca{}, despite being trained on fewer samples than the supervised baselines.
Furthermore, we outperform the Diff3F \emph{base features} on both \textit{SHREC} datasets in terms of accuracy.
While Diff3F achieves a higher accuracy at 1\% threshold in \dtosca{}, \refFig{accuracy_graph} shows that the accuracy of our model is higher for thresholds above $\approx2$\%.
This suggests that our method excels in the removal of outliers that can be caused by mismatched components.
Additionally, adapting our features to Functional maps~\cite{ovsjanikov2012functional} (\emph{+ FM}) distributes the error towards a lower mean at the cost of accuracy, and maintains the beneficial comparison to Diff3F+FM.

\refFig{qualitative_comparison} shows that Diff3F struggles to separate intraclass instances such as left and right legs.
In contrast, the results confirm the effectiveness of our contrastive loss in mitigating this issue.
We observe the same behavior for \dshrecanimals{} in \refFig{tosca_shapenet_examples} (\emph{top}), which contains highly diverse animal shapes. 
Furthermore, our method generally produces visually smoother results (see \refSupp{qualitative_shrec20} and the supplementary videos on the project website).

\begin{table}
\centering

\caption{Comparison of our 3D correspondence matching to prior works 3DC (3D-CODED~\cite{groueix20183d}), DPC~\cite{lang2021dpc}, SEN (SE-OrNet~\cite{deng2023se}), and Diff3F \cite{dutt2024diffusion}. 
†) Numbers originate from \cite{dutt2024diffusion}, *) Experiments were replicated, x) Omitted due to non-manifold meshes, + FM) Semantic features combined with Functional Maps. 
Accuracy is for the commonly used $\epsilon = 1$\,\%.
The per-column best results are bold and the second-to-best results are underlined.
}
\small
\begin{tabular}{lrrrr}
\toprule
 &   & SHREC’19 & TOSCA & SHREC’20 \\
\midrule
\multirow[c]{2}{*}{3DC†} & err \textdownarrow & 8.10 & 19.20 & - \\
 & acc \textuparrow & 2.10 & 0.50 & - \\
\cline{1-5}
\multirow[c]{2}{*}{DPC†} & err \textdownarrow & 6.26 & \underline{3.74} & \underline{2.13} \\
 & acc \textuparrow & 17.40 & 30.79 & 31.08 \\
\cline{1-5}
\multirow[c]{2}{*}{SEN†} & err \textdownarrow & 4.56 & 4.32 & \textbf{1.00} \\
 & acc \textuparrow & 21.41 & \textbf{33.25} & 31.70 \\
\hline
\hline
\multirow[c]{2}{*}{Diff3F*} & err \textdownarrow & 1.69\footnotesize{±1.44} & 4.51\footnotesize{±5.48} & 5.34\footnotesize{±10.22} \\
 & acc \textuparrow & \underline{26.25\footnotesize{±9.30}} & \underline{31.00\footnotesize{±15.73}} & \underline{69.50\footnotesize{±24.99}} \\
\cline{1-5}

\multirow[c]{2}{*}{Diff3F + FM*} & err \textdownarrow & 1.51\footnotesize{±1.65} & x & 4.44\footnotesize{±7.87} \\
 & acc \textuparrow & 21.71\footnotesize{±7.12} & x & 58.03\footnotesize{±25.94} \\
\cline{1-5}

\multirow[c]{2}{*}{Ours} & err \textdownarrow & \underline{0.43\footnotesize{±0.76}} & \textbf{1.65\footnotesize{±2.15}} & 3.89\footnotesize{±8.90} \\
 & acc \textuparrow & \textbf{28.78\footnotesize{±9.30}} & 29.35\footnotesize{±14.53} & \textbf{73.97\footnotesize{±26.47}} \\
\cline{1-5}

\multirow[c]{2}{*}{Ours + FM} & err \textdownarrow & \textbf{0.24\footnotesize{±0.64}} & x & 3.54\footnotesize{±7.59} \\
 & acc \textuparrow & 24.83\footnotesize{±6.80} & x & 63.61\footnotesize{±24.34} \\
\cline{1-5}

\end{tabular}

\label{tbl:results_diff3f}
\end{table}

\paragraph{Other Shapes}
Our method is applicable beyond humanoid and animal shapes, which we show by training two additional encoders for a subset of 50 chairs and 50 airplanes from ShapeNet~\cite{chang2015shapenet}.
Here, we uniformly resample the mesh vertices for a better surface coverage~\cite{wang2022dual}.

For unseen shapes in \refFig{tosca_shapenet_examples}, our \emph{surface-aware features} again better distinguish same-class instances such as chair legs and airplane wings, supporting a wider applicability of our methodology. More examples are shown in \refSupp{qualitative_shapenet} and the supplementary videos.

\subsection{Ablations}
\label{sec:ablations}
We motivate our design choices by ablation on various parts of our method in \refTbl{ablations} following the setup of \refSec{exp_correspondences}.

\paragraph{Choice of Angular Space}
We demonstrate the effectiveness of our hyperspherical embedding by replacing our contrastive loss $\mathcal{L}_c$ (\refEq{contrastive_loss}) with three different options inspired by related work (see \refSupp{ablations_baselines}).
First, the Relative Geodesic Loss (\textit{RGL})~\cite{jiang2023neural} optimizes relative distances in a Euclidean embedding.
Similarly, the Naive Geodesic Loss (\textit{NGL}) minimizes absolute distances. 
Finally, the Geometrical Similarity Loss (\textit{GSL})~ \cite{chen2025dvmatcherdeformationbasednonrigidpoint} enforces similarity of feature and surface distances in a local neighborhood.
We remove feature and geodesic normalization wherever absolute magnitude needs to be learned.
In \refTbl{ablations} (top), we observe that, except for correspondence accuracy for TOSCA, our contrastive loss $\mathcal{L}_c$ outperforms all of the alternatives in the correspondence matching task.

\paragraph{Contrastive and Reconstruction Loss}
In \refTbl{ablations} (bottom), we individually assess our two losses. 
We see that the performance with only the reconstruction loss $\mathcal{L}_r$ is close to Diff3F.
This indicates that the gain in performance does not originate predominantly from a smaller embedding space or from access to training data.
Similarly, the contrastive loss $\mathcal{L}_c$ alone results in an accuracy drop compared to our full model.
This justifies our autoencoder approach with both losses playing an import role.
Ablations on the number of anchors can be found in\refSupp{ablations_baselines}.

\begin{table}[]

\caption{
Ablation on our method.
\emph{Above the bar:} 
Ablation on alternative losses inspired by related work~\cite{chen2025dvmatcherdeformationbasednonrigidpoint,jiang2023neural} compared to the unmodified Diff3F features.
\emph{Below the bar:}
Our full method compared to its reduced variant omitting losses $\mathcal{L}_c$ or $\mathcal{L}_r$. The per-column best results are bold and the second-to-best results are underlined.
}

\small
\centering

\begin{tabular}{lrrrr}
\toprule
 &   & SHREC’19 & TOSCA & SHREC’20 \\
\midrule
\multirow[c]{2}{*}{RGL} & err \textdownarrow & 0.80\footnotesize{±1.08} & 2.74\footnotesize{±2.53} & 5.29\footnotesize{±9.85} \\
 & acc \textuparrow & 20.16\footnotesize{±10.03} & 16.53\footnotesize{±10.41} & 55.23\footnotesize{±21.93} \\
\cline{1-5}
\multirow[c]{2}{*}{NGL} & err \textdownarrow & 0.54\footnotesize{±0.90} & 2.11\footnotesize{±2.02} & 4.86\footnotesize{±9.49} \\
 & acc \textuparrow & 18.84\footnotesize{±9.59} & 18.86\footnotesize{±11.50} & 58.97\footnotesize{±23.12} \\
\cline{1-5}
\multirow[c]{2}{*}{GSL} & err \textdownarrow & 1.72\footnotesize{±1.45} & 4.17\footnotesize{±5.21} & 4.34\footnotesize{±9.23} \\
 & acc \textuparrow & \underline{26.89\footnotesize{±9.07}} & 29.77\footnotesize{±14.92} & \underline{73.39\footnotesize{±26.31}} \\
\cline{1-5}
\multirow[c]{2}{*}{Diff3F} & err \textdownarrow & 1.69\footnotesize{±1.44} & 4.51\footnotesize{±5.48} & 5.34\footnotesize{±10.22} \\
 & acc \textuparrow & 26.25\footnotesize{±9.30} & \textbf{31.00\footnotesize{±15.73}} & 69.50\footnotesize{±24.99} \\
\hline\hline

\multirow[c]{2}{*}{only $\mathcal{L}_r$} & err \textdownarrow & 1.65\footnotesize{±1.44} & 4.70\footnotesize{±5.64} & 4.87\footnotesize{±9.38} \\
 & acc \textuparrow & 26.53\footnotesize{±9.19} & \underline{30.27\footnotesize{±15.17}} & 72.94\footnotesize{±26.21} \\
\cline{1-5}
\multirow[c]{2}{*}{only $\mathcal{L}_c$} & err \textdownarrow & \textbf{0.38\footnotesize{±0.61}} & \underline{1.67\footnotesize{±2.29}} & \underline{4.30\footnotesize{±9.31}} \\
 & acc \textuparrow & 26.21\footnotesize{±8.78} & 25.58\footnotesize{±13.88} & 70.08\footnotesize{±25.17} \\
\cline{1-5}
\multirow[c]{2}{*}{Ours} & err \textdownarrow & \underline{0.43\footnotesize{±0.76}} & \textbf{1.65\footnotesize{±2.15}} & \textbf{3.89\footnotesize{±8.90}} \\
 & acc \textuparrow & \textbf{28.78\footnotesize{±9.30}} & 29.35\footnotesize{±14.53} & \textbf{73.97\footnotesize{±26.47}} \\
\cline{1-5}

\end{tabular}

\label{tbl:ablations}
\end{table}
\section{Applications}
\label{sec:applications}
We present additional downstream tasks that benefit from our \emph{surface-aware features} learned in \refSec{experiments}.

\subsection{Instance-based Part Segmentation}
\label{sec:part_segmentation}
\begin{figure}
    \centering
    \includegraphics[width=\linewidth]{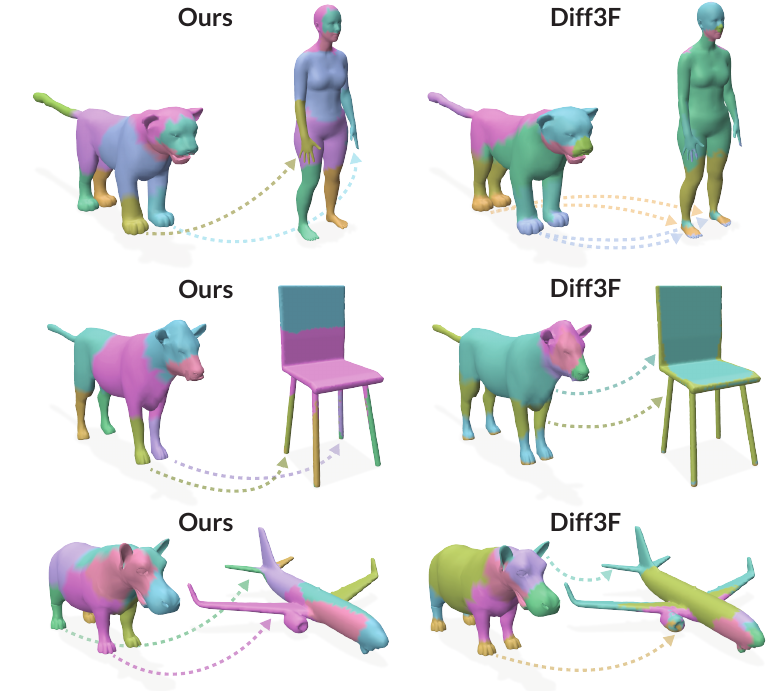}
    \caption{In the top row, 10 k-means cluster centers from the big cat were used to segment the human. 
    In the bottom two rows, 8 k-means cluster centers from the animals were used to segment the chairs and airplanes with a shared encoder.
    Unlike Diff3F, our method successfully separates all limbs for a plausible mapping from animal limbs to human limbs, chair legs, or airplane wings.
    }
    \label{fig:transfer_centroids}
\end{figure}

Following the prior work~\cite{dutt2024diffusion}, we segment a target shape by clustering features around centroids from K-means clustering of source-shape features.
In the top row of \refFig{transfer_centroids}, we demonstrate a transfer from a big cat to a human and see that, unlike the Diff3F features, our \emph{surface-aware features} disambiguate the limbs.
In the bottom two rows, we repeat this experiment with a shared encoder trained on human, animals, and a subset of \textit{ShapeNet} (see \refSupp{shared_encoder}) where a true mapping cannot be defined but our method finds reasonable analogies between the classes.

\begin{figure*}[!h]
    \centering
    \includegraphics[width=\textwidth]{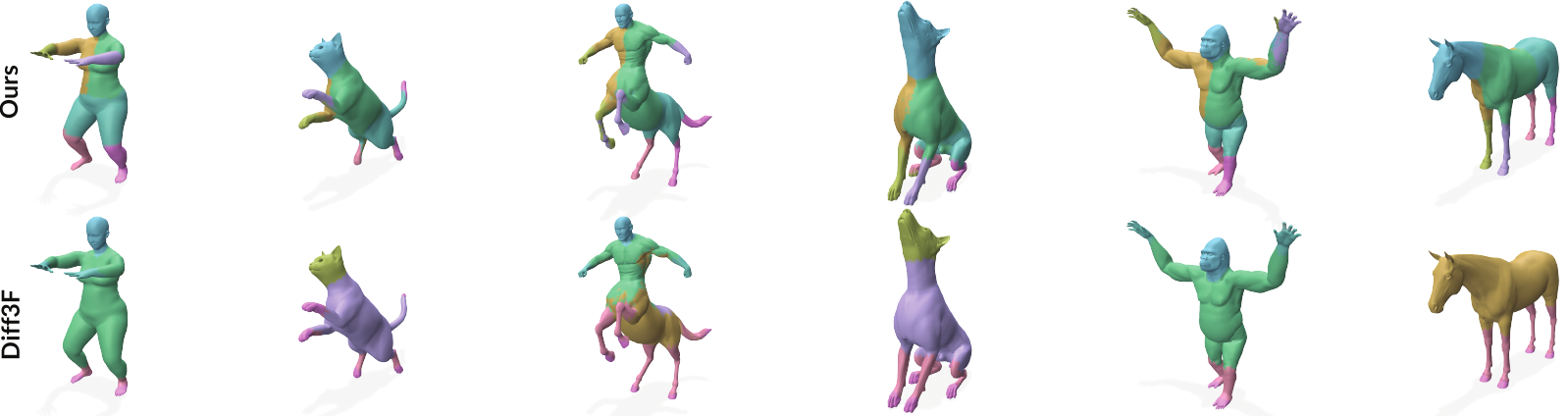}
    \caption{Results when clustering features across all samples in \damass{} and \dtosca{}. Our method implicitly aligns semantically-related regions (shown as the same colors) across diverse 3D shapes in a self-supervised manner (the top row). Diff3F produces inconsistent labeling across different shape categories as well as lack of separability between instanced components such as individual limbs (the bottom row).
    }
    \label{fig:cluster_comparison}
\end{figure*}

\begin{figure*}[!h]
    \centering
    \includegraphics[width=0.89\textwidth]{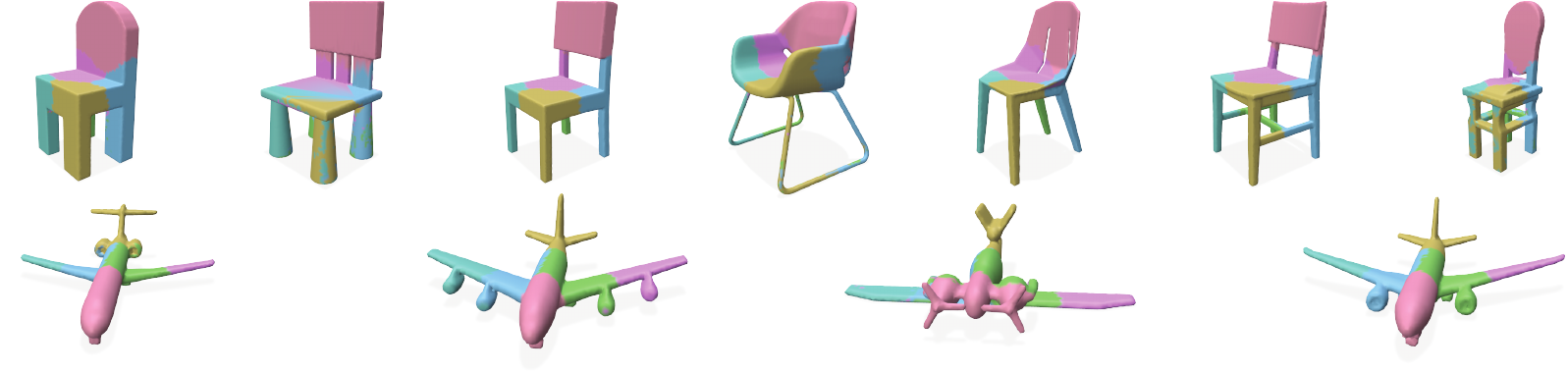}
    \caption{Results when clustering the \emph{surface-aware features} across chairs (top row) and airplanes (bottom row) from ShapeNet \cite{chang2015shapenet}. 
    Note, that while we use a single shared encoder for all humanoid and animal shapes, we train a separate encoder for each ShapeNet class due to the large domain gap. 
    }
    \label{fig:plane_chair_clustering}
\end{figure*}

In \refFig{cluster_comparison}, we repeat this with centroids obtained jointly from all \damass{} and \dtosca{} samples. 
In contrast to Diff3F, our method successfully matches features across diverse shapes, which demonstrates our embedding's capability of many-to-many shape matching without any additional pairwise optimization.
Finally, we show similar results for chairs and airplanes in \refFig{plane_chair_clustering}.

\begin{figure}
    \centering
    \includegraphics[width=\linewidth]{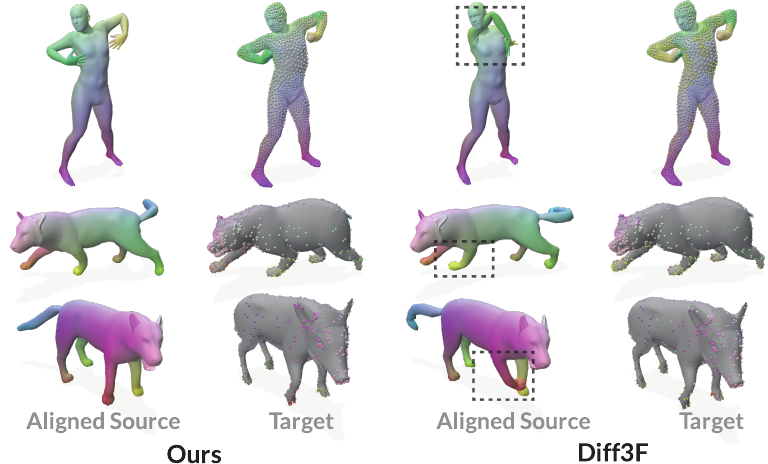}
    \caption{Pose alignment of a source shape (color) into the pose of a target (gray).
    The boxes highlight challenging areas handled well by our method.
    For humans, we densely fit all the target vertices, while for animals, we only fit $5$\,\% of the vertices as highlighted.}
    \label{fig:pose_alignment}
\end{figure}

\subsection{Pose Alignment}
\label{sec:pose_alignment}
Our \emph{surface-aware features} are also useful for pose alignment of a kinematic model to another 3D shape. 
To this end, we establish point correspondences between shape pairs as in \refSec{exp_correspondences} and optimize the kinematic pose parameters to minimize point-to-point distances (see \refSupp{pose_alignment}).

In \refFig{pose_alignment}, we align \damass{} to \dshrechumans{}, and \dsmalours{} to DeformingThings4D~\cite{li20214dcomplete} animals.
Benefiting from the robust instance separation, our method produces poses closer to the targets for both dense and sparse correspondences.
See our video for a 3D shape animation obtained by aligning to a target shape sequence.

\subsection{Texturing}
\begin{figure*}[]
    \centering
    \includegraphics[width=\textwidth]{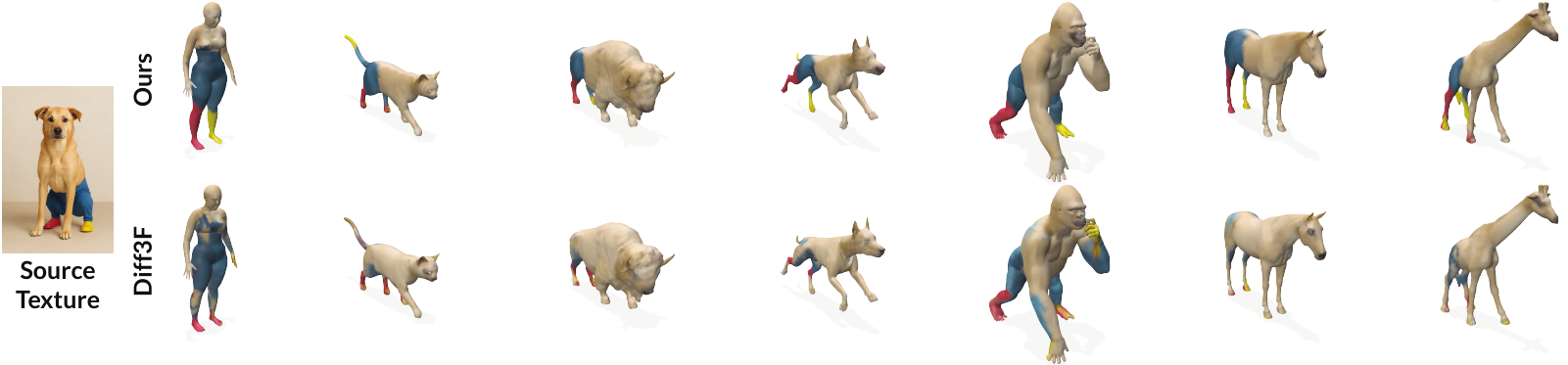}
    \caption{Texturing of 3D meshes from \damass{} and \dtosca{}, based on a 2D image generated with ChatGPT. The appearance is transferred by establishing correspondence between the image features and 3D mesh features. In contrast to Diff3F, our \emph{surface-aware features} represent the input image more faithfully.}
    \label{fig:2d_texturing}
\end{figure*}

\begin{figure*}[!h]
    \centering
    \includegraphics[width=\textwidth]{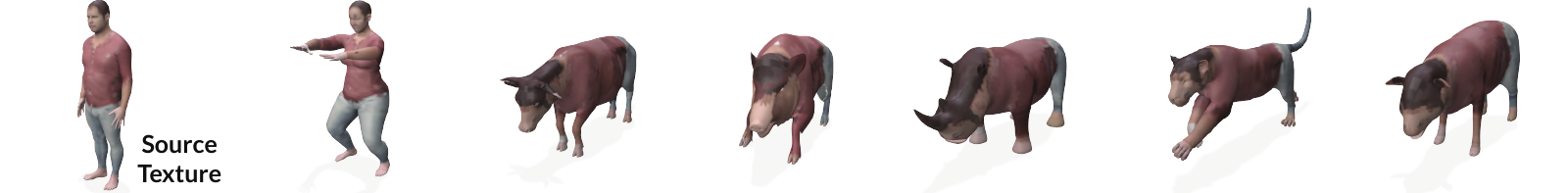}
    \caption{Texturing of 3D meshes based on a source 3D mesh. The appearance is transferred based on correspondences established by combining our \emph{surface-aware features} with Functional Maps.
    The data originate from \damass{}, SMPLitex~\cite{casas2023smplitex}, and DeformingThings4D \cite{li20214dcomplete}.}
    \label{fig:3d_texturing}
\end{figure*}

Since the \emph{base features} are obtained from image models (see \refSec{preliminaries}) and our pointwise encoder can process points sampled from a mesh as easily as pixels sampled from an image, we can establish correspondences between a 2D image and a 3D mesh. 
We demonstrate this by texturing 3D meshes from a masked target image
and individually assign each vertex a color from the image pixel that maximizes the mutual feature similarity (\refEq{cosine_similarity}) (see \refFig{2d_texturing} and \refSupp{texturing}).
We observe that our features produce a more coherent mapping leading to a better preservation of the source appearance when compared to the Diff3F \emph{base features}. 
In \refFig{3d_texturing}, we further show that textures can also be effectively transferred between two 3D shapes using a combination our \emph{surface-aware features} with Functional Maps like in \refSec{exp_correspondences}.

\section{Discussion}
\label{sec:discussion}

\paragraph{Limitations and Future Work}
Our method inherits limitations connected to the extraction of the \emph{base features}.
Specifically, the extraction of Diff3F features~\cite{dutt2024diffusion} takes several minutes per mesh and its vision model is sensitive to rendering artifacts or upside-down mesh orientations.
We expect that advances in rendering of point representations increase the applicability across representations~\cite{kerbl20233d}.
Furthermore, our method cannot establish a consistent partitioning for objects that are both geometrically and semantically isotropic (e.g., a round table). 
Hence, while our embedding separates human legs following the body's notion of front and rear, it cannot do so for table legs.
However, this is not an issue for applications such as shape morphing~\cite{sun2024srif}.
Lastly, our method relies on consistency of geodesic distances between semantically distinct parts, and therefore it will be affected by geodesic shortcuts for partially blended parts in noisy 3D reconstructions (e.g., touching hands).

Beyond 3D alignment, our methodology could inspire 3D-to-2D pose estimation~\cite{kanazawa2018end, peng2019pvnet}, articulated 3D reconstruction~\cite{yao2022lassie,uzolas2023template}, automated rigging~\cite{xu2020rignet} or 2D-to-3D uplifting~\cite{poole2022dreamfusion, liu2023zero}, where our features could support more view-consistent representations.
Finally, an interesting topic for future research is the development of foundational features using massive datasets, such as Objaverse~\cite{deitke2023objaverse}.

\paragraph{Conclusion}
We have introduced novel \emph{surface-aware features} for 3D shape matching that disambiguate intra-class instances among semantic features derived from pre-trained 2D vision models.
Our descriptors have proven effective in distinguishing instances of the same semantic class and they generalize even when trained on a limited number of 3D shapes.
Furthermore, our contrastive loss facilitates easy integration in future unsupervised methods which reduces data labeling effort.
Consequently, our method is a promising building block toward adapting pre-trained 2D models to 3D tasks.

\begin{acks}
This work was partially supported by the Convergence AI Immersive Tech Lab at TU Delft.
\end{acks}

\bibliographystyle{ACM-Reference-Format}
\bibliography{main}


\begin{thebibliography}{104}


\ifx \showCODEN    \undefined \def \showCODEN     #1{\unskip}     \fi
\ifx \showISBNx    \undefined \def \showISBNx     #1{\unskip}     \fi
\ifx \showISBNxiii \undefined \def \showISBNxiii  #1{\unskip}     \fi
\ifx \showISSN     \undefined \def \showISSN      #1{\unskip}     \fi
\ifx \showLCCN     \undefined \def \showLCCN      #1{\unskip}     \fi
\ifx \shownote     \undefined \def \shownote      #1{#1}          \fi
\ifx \showarticletitle \undefined \def \showarticletitle #1{#1}   \fi
\ifx \showURL      \undefined \def \showURL       {\relax}        \fi
\providecommand\bibfield[2]{#2}
\providecommand\bibinfo[2]{#2}
\providecommand\natexlab[1]{#1}
\providecommand\showeprint[2][]{arXiv:#2}

\bibitem[Abdelreheem et~al\mbox{.}(2023)]%
        {abdelreheem2023zero}
\bibfield{author}{\bibinfo{person}{Ahmed Abdelreheem}, \bibinfo{person}{Abdelrahman Eldesokey}, \bibinfo{person}{Maks Ovsjanikov}, {and} \bibinfo{person}{Peter Wonka}.} \bibinfo{year}{2023}\natexlab{}.
\newblock \showarticletitle{Zero-shot 3d shape correspondence}. In \bibinfo{booktitle}{\emph{SIGGRAPH Asia 2023 Conference Papers}}. \bibinfo{pages}{1--11}.
\newblock


\bibitem[Aigerman et~al\mbox{.}(2022)]%
        {aigerman2022neural}
\bibfield{author}{\bibinfo{person}{Noam Aigerman}, \bibinfo{person}{Kunal Gupta}, \bibinfo{person}{Vladimir~G Kim}, \bibinfo{person}{Siddhartha Chaudhuri}, \bibinfo{person}{Jun Saito}, {and} \bibinfo{person}{Thibault Groueix}.} \bibinfo{year}{2022}\natexlab{}.
\newblock \showarticletitle{Neural jacobian fields: learning intrinsic mappings of arbitrary meshes}.
\newblock \bibinfo{journal}{\emph{ACM Transactions on Graphics (TOG)}} \bibinfo{volume}{41}, \bibinfo{number}{4} (\bibinfo{year}{2022}), \bibinfo{pages}{1--17}.
\newblock


\bibitem[Amir et~al\mbox{.}(2021)]%
        {amir2021deep}
\bibfield{author}{\bibinfo{person}{Shir Amir}, \bibinfo{person}{Yossi Gandelsman}, \bibinfo{person}{Shai Bagon}, {and} \bibinfo{person}{Tali Dekel}.} \bibinfo{year}{2021}\natexlab{}.
\newblock \showarticletitle{Deep vit features as dense visual descriptors}.
\newblock \bibinfo{journal}{\emph{arXiv preprint arXiv:2112.05814}} \bibinfo{volume}{2}, \bibinfo{number}{3} (\bibinfo{year}{2021}), \bibinfo{pages}{4}.
\newblock


\bibitem[Ansel et~al\mbox{.}(2024)]%
        {ansel2024pytorch}
\bibfield{author}{\bibinfo{person}{Jason Ansel}, \bibinfo{person}{Edward Yang}, \bibinfo{person}{Horace He}, \bibinfo{person}{Natalia Gimelshein}, \bibinfo{person}{Animesh Jain}, \bibinfo{person}{Michael Voznesensky}, \bibinfo{person}{Bin Bao}, \bibinfo{person}{Peter Bell}, \bibinfo{person}{David Berard}, \bibinfo{person}{Evgeni Burovski}, {et~al\mbox{.}}} \bibinfo{year}{2024}\natexlab{}.
\newblock \showarticletitle{Pytorch 2: Faster machine learning through dynamic python bytecode transformation and graph compilation}. In \bibinfo{booktitle}{\emph{Proceedings of the 29th ACM International Conference on Architectural Support for Programming Languages and Operating Systems, Volume 2}}. \bibinfo{pages}{929--947}.
\newblock


\bibitem[Attaiki et~al\mbox{.}(2023)]%
        {attaiki2023generalizable}
\bibfield{author}{\bibinfo{person}{Souhaib Attaiki}, \bibinfo{person}{Lei Li}, {and} \bibinfo{person}{Maks Ovsjanikov}.} \bibinfo{year}{2023}\natexlab{}.
\newblock \showarticletitle{Generalizable local feature pre-training for deformable shape analysis}. In \bibinfo{booktitle}{\emph{Proceedings of the IEEE/CVF Conference on Computer Vision and Pattern Recognition}}. \bibinfo{pages}{13650--13661}.
\newblock


\bibitem[Aubry et~al\mbox{.}(2011)]%
        {aubry2011wave}
\bibfield{author}{\bibinfo{person}{Mathieu Aubry}, \bibinfo{person}{Ulrich Schlickewei}, {and} \bibinfo{person}{Daniel Cremers}.} \bibinfo{year}{2011}\natexlab{}.
\newblock \showarticletitle{The wave kernel signature: A quantum mechanical approach to shape analysis}. In \bibinfo{booktitle}{\emph{2011 IEEE international conference on computer vision workshops (ICCV workshops)}}. IEEE, \bibinfo{pages}{1626--1633}.
\newblock


\bibitem[Ba et~al\mbox{.}(2016)]%
        {ba2016layer}
\bibfield{author}{\bibinfo{person}{Jimmy~Lei Ba}, \bibinfo{person}{Jamie~Ryan Kiros}, {and} \bibinfo{person}{Geoffrey~E Hinton}.} \bibinfo{year}{2016}\natexlab{}.
\newblock \showarticletitle{Layer normalization}.
\newblock \bibinfo{journal}{\emph{arXiv preprint arXiv:1607.06450}} (\bibinfo{year}{2016}).
\newblock


\bibitem[Besl and McKay(1992)]%
        {besl1992method}
\bibfield{author}{\bibinfo{person}{Paul~J Besl} {and} \bibinfo{person}{Neil~D McKay}.} \bibinfo{year}{1992}\natexlab{}.
\newblock \showarticletitle{Method for registration of 3-D shapes}. In \bibinfo{booktitle}{\emph{Sensor fusion IV: control paradigms and data structures}}, Vol.~\bibinfo{volume}{1611}. Spie, \bibinfo{pages}{586--606}.
\newblock


\bibitem[Biggs et~al\mbox{.}(2018)]%
        {biggs2018creatures}
\bibfield{author}{\bibinfo{person}{Benjamin Biggs}, \bibinfo{person}{Thomas Roddick}, \bibinfo{person}{Andrew Fitzgibbon}, {and} \bibinfo{person}{Roberto Cipolla}.} \bibinfo{year}{2018}\natexlab{}.
\newblock \showarticletitle{{C}reatures great and {SMAL}: {R}ecovering the shape and motion of animals from video}. In \bibinfo{booktitle}{\emph{ACCV}}.
\newblock


\bibitem[Bronstein et~al\mbox{.}(2008)]%
        {bronstein2008numerical}
\bibfield{author}{\bibinfo{person}{Alexander~M Bronstein}, \bibinfo{person}{Michael~M Bronstein}, {and} \bibinfo{person}{Ron Kimmel}.} \bibinfo{year}{2008}\natexlab{}.
\newblock \bibinfo{booktitle}{\emph{Numerical geometry of non-rigid shapes}}.
\newblock \bibinfo{publisher}{Springer Science \& Business Media}.
\newblock


\bibitem[Brown and Rusinkiewicz(2007)]%
        {brown2007global}
\bibfield{author}{\bibinfo{person}{Benedict~J Brown} {and} \bibinfo{person}{Szymon Rusinkiewicz}.} \bibinfo{year}{2007}\natexlab{}.
\newblock \showarticletitle{Global non-rigid alignment of 3-D scans}.
\newblock \bibinfo{journal}{\emph{ACM Transactions on Graphics (TOG)}} \bibinfo{volume}{26}, \bibinfo{number}{3} (\bibinfo{year}{2007}), \bibinfo{pages}{21--es}.
\newblock


\bibitem[Caron et~al\mbox{.}(2018)]%
        {caron2018deep}
\bibfield{author}{\bibinfo{person}{Mathilde Caron}, \bibinfo{person}{Piotr Bojanowski}, \bibinfo{person}{Armand Joulin}, {and} \bibinfo{person}{Matthijs Douze}.} \bibinfo{year}{2018}\natexlab{}.
\newblock \showarticletitle{Deep clustering for unsupervised learning of visual features}. In \bibinfo{booktitle}{\emph{Proceedings of the European conference on computer vision (ECCV)}}. \bibinfo{pages}{132--149}.
\newblock


\bibitem[Caron et~al\mbox{.}(2020)]%
        {caron2020unsupervised}
\bibfield{author}{\bibinfo{person}{Mathilde Caron}, \bibinfo{person}{Ishan Misra}, \bibinfo{person}{Julien Mairal}, \bibinfo{person}{Priya Goyal}, \bibinfo{person}{Piotr Bojanowski}, {and} \bibinfo{person}{Armand Joulin}.} \bibinfo{year}{2020}\natexlab{}.
\newblock \showarticletitle{Unsupervised learning of visual features by contrasting cluster assignments}.
\newblock \bibinfo{journal}{\emph{Advances in neural information processing systems}}  \bibinfo{volume}{33} (\bibinfo{year}{2020}), \bibinfo{pages}{9912--9924}.
\newblock


\bibitem[Caron et~al\mbox{.}(2021)]%
        {caron2021emerging}
\bibfield{author}{\bibinfo{person}{Mathilde Caron}, \bibinfo{person}{Hugo Touvron}, \bibinfo{person}{Ishan Misra}, \bibinfo{person}{Herv{\'e} J{\'e}gou}, \bibinfo{person}{Julien Mairal}, \bibinfo{person}{Piotr Bojanowski}, {and} \bibinfo{person}{Armand Joulin}.} \bibinfo{year}{2021}\natexlab{}.
\newblock \showarticletitle{Emerging properties in self-supervised vision transformers}. In \bibinfo{booktitle}{\emph{Proceedings of the IEEE/CVF international conference on computer vision}}. \bibinfo{pages}{9650--9660}.
\newblock


\bibitem[Casas and Comino-Trinidad(2023)]%
        {casas2023smplitex}
\bibfield{author}{\bibinfo{person}{Dan Casas} {and} \bibinfo{person}{Marc Comino-Trinidad}.} \bibinfo{year}{2023}\natexlab{}.
\newblock \showarticletitle{{SMPLitex: A Generative Model and Dataset for 3D Human Texture Estimation from Single Image}}. In \bibinfo{booktitle}{\emph{British Machine Vision Conference (BMVC)}}.
\newblock


\bibitem[Chang et~al\mbox{.}(2015)]%
        {chang2015shapenet}
\bibfield{author}{\bibinfo{person}{Angel~X Chang}, \bibinfo{person}{Thomas Funkhouser}, \bibinfo{person}{Leonidas Guibas}, \bibinfo{person}{Pat Hanrahan}, \bibinfo{person}{Qixing Huang}, \bibinfo{person}{Zimo Li}, \bibinfo{person}{Silvio Savarese}, \bibinfo{person}{Manolis Savva}, \bibinfo{person}{Shuran Song}, \bibinfo{person}{Hao Su}, {et~al\mbox{.}}} \bibinfo{year}{2015}\natexlab{}.
\newblock \showarticletitle{Shapenet: An information-rich 3d model repository}.
\newblock \bibinfo{journal}{\emph{arXiv preprint arXiv:1512.03012}} (\bibinfo{year}{2015}).
\newblock


\bibitem[Chen et~al\mbox{.}(2020)]%
        {chen2020simple}
\bibfield{author}{\bibinfo{person}{Ting Chen}, \bibinfo{person}{Simon Kornblith}, \bibinfo{person}{Mohammad Norouzi}, {and} \bibinfo{person}{Geoffrey Hinton}.} \bibinfo{year}{2020}\natexlab{}.
\newblock \showarticletitle{A simple framework for contrastive learning of visual representations}. In \bibinfo{booktitle}{\emph{International conference on machine learning}}. PmLR, \bibinfo{pages}{1597--1607}.
\newblock


\bibitem[Chen et~al\mbox{.}(2025)]%
        {chen2025dvmatcherdeformationbasednonrigidpoint}
\bibfield{author}{\bibinfo{person}{Zhangquan Chen}, \bibinfo{person}{Puhua Jiang}, {and} \bibinfo{person}{Ruqi Huang}.} \bibinfo{year}{2025}\natexlab{}.
\newblock \showarticletitle{DV-Matcher: Deformation-based Non-Rigid Point Cloud Matching Guided by Pre-trained Visual Features}.
\newblock \bibinfo{journal}{\emph{arXiv preprint arXiv:2408.08568v2}} (\bibinfo{year}{2025}).
\newblock


\bibitem[Chopra et~al\mbox{.}(2005)]%
        {chopra2005learning}
\bibfield{author}{\bibinfo{person}{Sumit Chopra}, \bibinfo{person}{Raia Hadsell}, {and} \bibinfo{person}{Yann LeCun}.} \bibinfo{year}{2005}\natexlab{}.
\newblock \showarticletitle{Learning a similarity metric discriminatively, with application to face verification}. In \bibinfo{booktitle}{\emph{2005 IEEE computer society conference on computer vision and pattern recognition (CVPR'05)}}, Vol.~\bibinfo{volume}{1}. IEEE, \bibinfo{pages}{539--546}.
\newblock


\bibitem[Corman et~al\mbox{.}(2014)]%
        {corman2014supervised}
\bibfield{author}{\bibinfo{person}{Etienne Corman}, \bibinfo{person}{Maks Ovsjanikov}, {and} \bibinfo{person}{Antonin Chambolle}.} \bibinfo{year}{2014}\natexlab{}.
\newblock \showarticletitle{Supervised descriptor learning for non-rigid shape matching}. In \bibinfo{booktitle}{\emph{European conference on computer vision}}. Springer, \bibinfo{pages}{283--298}.
\newblock


\bibitem[Crane et~al\mbox{.}(2017)]%
        {crane2017heat}
\bibfield{author}{\bibinfo{person}{Keenan Crane}, \bibinfo{person}{Clarisse Weischedel}, {and} \bibinfo{person}{Max Wardetzky}.} \bibinfo{year}{2017}\natexlab{}.
\newblock \showarticletitle{The heat method for distance computation}.
\newblock \bibinfo{journal}{\emph{Commun. ACM}} \bibinfo{volume}{60}, \bibinfo{number}{11} (\bibinfo{year}{2017}), \bibinfo{pages}{90--99}.
\newblock


\bibitem[Deitke et~al\mbox{.}(2023)]%
        {deitke2023objaverse}
\bibfield{author}{\bibinfo{person}{Matt Deitke}, \bibinfo{person}{Ruoshi Liu}, \bibinfo{person}{Matthew Wallingford}, \bibinfo{person}{Huong Ngo}, \bibinfo{person}{Oscar Michel}, \bibinfo{person}{Aditya Kusupati}, \bibinfo{person}{Alan Fan}, \bibinfo{person}{Christian Laforte}, \bibinfo{person}{Vikram Voleti}, \bibinfo{person}{Samir~Yitzhak Gadre}, {et~al\mbox{.}}} \bibinfo{year}{2023}\natexlab{}.
\newblock \showarticletitle{Objaverse-xl: A universe of 10m+ 3d objects}.
\newblock \bibinfo{journal}{\emph{Advances in Neural Information Processing Systems}}  \bibinfo{volume}{36} (\bibinfo{year}{2023}), \bibinfo{pages}{35799--35813}.
\newblock


\bibitem[Deng et~al\mbox{.}(2018a)]%
        {deng2018ppffoldnet}
\bibfield{author}{\bibinfo{person}{Haowen Deng}, \bibinfo{person}{Tolga Birdal}, {and} \bibinfo{person}{Slobodan Ilic}.} \bibinfo{year}{2018}\natexlab{a}.
\newblock \showarticletitle{Ppf-foldnet: Unsupervised learning of rotation invariant 3d local descriptors}. In \bibinfo{booktitle}{\emph{Proceedings of the European conference on computer vision (ECCV)}}. \bibinfo{pages}{602--618}.
\newblock


\bibitem[Deng et~al\mbox{.}(2018b)]%
        {deng2018ppfnet}
\bibfield{author}{\bibinfo{person}{Haowen Deng}, \bibinfo{person}{Tolga Birdal}, {and} \bibinfo{person}{Slobodan Ilic}.} \bibinfo{year}{2018}\natexlab{b}.
\newblock \showarticletitle{Ppfnet: Global context aware local features for robust 3d point matching}. In \bibinfo{booktitle}{\emph{Proceedings of the IEEE conference on computer vision and pattern recognition}}. \bibinfo{pages}{195--205}.
\newblock


\bibitem[Deng et~al\mbox{.}(2023)]%
        {deng2023se}
\bibfield{author}{\bibinfo{person}{Jiacheng Deng}, \bibinfo{person}{Chuxin Wang}, \bibinfo{person}{Jiahao Lu}, \bibinfo{person}{Jianfeng He}, \bibinfo{person}{Tianzhu Zhang}, \bibinfo{person}{Jiyang Yu}, {and} \bibinfo{person}{Zhe Zhang}.} \bibinfo{year}{2023}\natexlab{}.
\newblock \showarticletitle{Se-ornet: Self-ensembling orientation-aware network for unsupervised point cloud shape correspondence}. In \bibinfo{booktitle}{\emph{Proceedings of the IEEE/CVF Conference on Computer Vision and Pattern Recognition}}. \bibinfo{pages}{5364--5373}.
\newblock


\bibitem[Deprelle et~al\mbox{.}(2019)]%
        {deprelle2019learning}
\bibfield{author}{\bibinfo{person}{Theo Deprelle}, \bibinfo{person}{Thibault Groueix}, \bibinfo{person}{Matthew Fisher}, \bibinfo{person}{Vladimir Kim}, \bibinfo{person}{Bryan Russell}, {and} \bibinfo{person}{Mathieu Aubry}.} \bibinfo{year}{2019}\natexlab{}.
\newblock \showarticletitle{Learning elementary structures for 3d shape generation and matching}.
\newblock \bibinfo{journal}{\emph{Advances in Neural Information Processing Systems}}  \bibinfo{volume}{32} (\bibinfo{year}{2019}).
\newblock


\bibitem[Donati et~al\mbox{.}(2020)]%
        {donati2020deep}
\bibfield{author}{\bibinfo{person}{Nicolas Donati}, \bibinfo{person}{Abhishek Sharma}, {and} \bibinfo{person}{Maks Ovsjanikov}.} \bibinfo{year}{2020}\natexlab{}.
\newblock \showarticletitle{Deep geometric functional maps: Robust feature learning for shape correspondence}. In \bibinfo{booktitle}{\emph{Proceedings of the IEEE/CVF Conference on Computer Vision and Pattern Recognition}}. \bibinfo{pages}{8592--8601}.
\newblock


\bibitem[Dutt et~al\mbox{.}(2024)]%
        {dutt2024diffusion}
\bibfield{author}{\bibinfo{person}{Niladri~Shekhar Dutt}, \bibinfo{person}{Sanjeev Muralikrishnan}, {and} \bibinfo{person}{Niloy~J Mitra}.} \bibinfo{year}{2024}\natexlab{}.
\newblock \showarticletitle{Diffusion 3d features (diff3f): Decorating untextured shapes with distilled semantic features}. In \bibinfo{booktitle}{\emph{Proceedings of the IEEE/CVF Conference on Computer Vision and Pattern Recognition}}. \bibinfo{pages}{4494--4504}.
\newblock


\bibitem[Dyke et~al\mbox{.}(2020)]%
        {dyke2020shrec}
\bibfield{author}{\bibinfo{person}{Roberto~M Dyke}, \bibinfo{person}{Yu-Kun Lai}, \bibinfo{person}{Paul~L Rosin}, \bibinfo{person}{Stefano Zappal{\`a}}, \bibinfo{person}{Seana Dykes}, \bibinfo{person}{Daoliang Guo}, \bibinfo{person}{Kun Li}, \bibinfo{person}{Riccardo Marin}, \bibinfo{person}{Simone Melzi}, {and} \bibinfo{person}{Jingyu Yang}.} \bibinfo{year}{2020}\natexlab{}.
\newblock \showarticletitle{SHREC’20: Shape correspondence with non-isometric deformations}.
\newblock \bibinfo{journal}{\emph{Computers \& Graphics}}  \bibinfo{volume}{92} (\bibinfo{year}{2020}), \bibinfo{pages}{28--43}.
\newblock


\bibitem[Eisenberger et~al\mbox{.}(2020a)]%
        {eisenberger2020smooth}
\bibfield{author}{\bibinfo{person}{Marvin Eisenberger}, \bibinfo{person}{Zorah Lahner}, {and} \bibinfo{person}{Daniel Cremers}.} \bibinfo{year}{2020}\natexlab{a}.
\newblock \showarticletitle{Smooth shells: Multi-scale shape registration with functional maps}. In \bibinfo{booktitle}{\emph{Proceedings of the IEEE/CVF Conference on Computer Vision and Pattern Recognition}}. \bibinfo{pages}{12265--12274}.
\newblock


\bibitem[Eisenberger et~al\mbox{.}(2020b)]%
        {eisenberger2020deep}
\bibfield{author}{\bibinfo{person}{Marvin Eisenberger}, \bibinfo{person}{Aysim Toker}, \bibinfo{person}{Laura Leal-Taix{\'e}}, {and} \bibinfo{person}{Daniel Cremers}.} \bibinfo{year}{2020}\natexlab{b}.
\newblock \showarticletitle{Deep shells: Unsupervised shape correspondence with optimal transport}.
\newblock \bibinfo{journal}{\emph{Advances in Neural information processing systems}}  \bibinfo{volume}{33} (\bibinfo{year}{2020}), \bibinfo{pages}{10491--10502}.
\newblock


\bibitem[Elbaz et~al\mbox{.}(2017)]%
        {elbaz20173d}
\bibfield{author}{\bibinfo{person}{Gil Elbaz}, \bibinfo{person}{Tamar Avraham}, {and} \bibinfo{person}{Anath Fischer}.} \bibinfo{year}{2017}\natexlab{}.
\newblock \showarticletitle{3D point cloud registration for localization using a deep neural network auto-encoder}. In \bibinfo{booktitle}{\emph{Proceedings of the IEEE conference on computer vision and pattern recognition}}. \bibinfo{pages}{4631--4640}.
\newblock


\bibitem[Elfwing et~al\mbox{.}(2018)]%
        {elfwing2018sigmoid}
\bibfield{author}{\bibinfo{person}{Stefan Elfwing}, \bibinfo{person}{Eiji Uchibe}, {and} \bibinfo{person}{Kenji Doya}.} \bibinfo{year}{2018}\natexlab{}.
\newblock \showarticletitle{Sigmoid-weighted linear units for neural network function approximation in reinforcement learning}.
\newblock \bibinfo{journal}{\emph{Neural networks}}  \bibinfo{volume}{107} (\bibinfo{year}{2018}), \bibinfo{pages}{3--11}.
\newblock


\bibitem[Fischer et~al\mbox{.}(2024)]%
        {fischer2024nerf}
\bibfield{author}{\bibinfo{person}{Michael Fischer}, \bibinfo{person}{Zhengqin Li}, \bibinfo{person}{Thu Nguyen-Phuoc}, \bibinfo{person}{Aljaz Bozic}, \bibinfo{person}{Zhao Dong}, \bibinfo{person}{Carl Marshall}, {and} \bibinfo{person}{Tobias Ritschel}.} \bibinfo{year}{2024}\natexlab{}.
\newblock \showarticletitle{NeRF Analogies: Example-Based Visual Attribute Transfer for NeRFs}. In \bibinfo{booktitle}{\emph{Proceedings of the IEEE/CVF Conference on Computer Vision and Pattern Recognition}}. \bibinfo{pages}{4640--4650}.
\newblock


\bibitem[Fischler and Bolles(1981)]%
        {fischler1981random}
\bibfield{author}{\bibinfo{person}{Martin~A Fischler} {and} \bibinfo{person}{Robert~C Bolles}.} \bibinfo{year}{1981}\natexlab{}.
\newblock \showarticletitle{Random sample consensus: a paradigm for model fitting with applications to image analysis and automated cartography}.
\newblock \bibinfo{journal}{\emph{Commun. ACM}} \bibinfo{volume}{24}, \bibinfo{number}{6} (\bibinfo{year}{1981}), \bibinfo{pages}{381--395}.
\newblock


\bibitem[Ginzburg and Raviv(2020)]%
        {ginzburg2020cyclic}
\bibfield{author}{\bibinfo{person}{Dvir Ginzburg} {and} \bibinfo{person}{Dan Raviv}.} \bibinfo{year}{2020}\natexlab{}.
\newblock \showarticletitle{Cyclic functional mapping: Self-supervised correspondence between non-isometric deformable shapes}. In \bibinfo{booktitle}{\emph{Computer Vision--ECCV 2020: 16th European Conference, Glasgow, UK, August 23--28, 2020, Proceedings, Part V 16}}. Springer, \bibinfo{pages}{36--52}.
\newblock


\bibitem[Gojcic et~al\mbox{.}(2019)]%
        {gojcic2019perfect}
\bibfield{author}{\bibinfo{person}{Zan Gojcic}, \bibinfo{person}{Caifa Zhou}, \bibinfo{person}{Jan~D Wegner}, {and} \bibinfo{person}{Andreas Wieser}.} \bibinfo{year}{2019}\natexlab{}.
\newblock \showarticletitle{The perfect match: 3d point cloud matching with smoothed densities}. In \bibinfo{booktitle}{\emph{Proceedings of the IEEE/CVF conference on computer vision and pattern recognition}}. \bibinfo{pages}{5545--5554}.
\newblock


\bibitem[Gong et~al\mbox{.}(2024)]%
        {gong2024zerokey}
\bibfield{author}{\bibinfo{person}{Bingchen Gong}, \bibinfo{person}{Diego Gomez}, \bibinfo{person}{Abdullah Hamdi}, \bibinfo{person}{Abdelrahman Eldesokey}, \bibinfo{person}{Ahmed Abdelreheem}, \bibinfo{person}{Peter Wonka}, {and} \bibinfo{person}{Maks Ovsjanikov}.} \bibinfo{year}{2024}\natexlab{}.
\newblock \showarticletitle{ZeroKey: Point-Level Reasoning and Zero-Shot 3D Keypoint Detection from Large Language Models}.
\newblock \bibinfo{journal}{\emph{arXiv preprint arXiv:2412.06292}} (\bibinfo{year}{2024}).
\newblock


\bibitem[Groueix et~al\mbox{.}(2018)]%
        {groueix20183d}
\bibfield{author}{\bibinfo{person}{Thibault Groueix}, \bibinfo{person}{Matthew Fisher}, \bibinfo{person}{Vladimir~G Kim}, \bibinfo{person}{Bryan~C Russell}, {and} \bibinfo{person}{Mathieu Aubry}.} \bibinfo{year}{2018}\natexlab{}.
\newblock \showarticletitle{3d-coded: 3d correspondences by deep deformation}. In \bibinfo{booktitle}{\emph{Proceedings of the european conference on computer vision (ECCV)}}. \bibinfo{pages}{230--246}.
\newblock


\bibitem[Guo et~al\mbox{.}(2015)]%
        {guo20153d}
\bibfield{author}{\bibinfo{person}{Kan Guo}, \bibinfo{person}{Dongqing Zou}, {and} \bibinfo{person}{Xiaowu Chen}.} \bibinfo{year}{2015}\natexlab{}.
\newblock \showarticletitle{3D mesh labeling via deep convolutional neural networks}.
\newblock \bibinfo{journal}{\emph{ACM Transactions on Graphics (TOG)}} \bibinfo{volume}{35}, \bibinfo{number}{1} (\bibinfo{year}{2015}), \bibinfo{pages}{1--12}.
\newblock


\bibitem[Hadsell et~al\mbox{.}(2006)]%
        {hadsell2006dimensionality}
\bibfield{author}{\bibinfo{person}{Raia Hadsell}, \bibinfo{person}{Sumit Chopra}, {and} \bibinfo{person}{Yann LeCun}.} \bibinfo{year}{2006}\natexlab{}.
\newblock \showarticletitle{Dimensionality reduction by learning an invariant mapping}. In \bibinfo{booktitle}{\emph{2006 IEEE computer society conference on computer vision and pattern recognition (CVPR'06)}}, Vol.~\bibinfo{volume}{2}. IEEE, \bibinfo{pages}{1735--1742}.
\newblock


\bibitem[Halimi et~al\mbox{.}(2019)]%
        {halimi2019unsupervised}
\bibfield{author}{\bibinfo{person}{Oshri Halimi}, \bibinfo{person}{Or Litany}, \bibinfo{person}{Emanuele Rodola}, \bibinfo{person}{Alex~M Bronstein}, {and} \bibinfo{person}{Ron Kimmel}.} \bibinfo{year}{2019}\natexlab{}.
\newblock \showarticletitle{Unsupervised learning of dense shape correspondence}. In \bibinfo{booktitle}{\emph{Proceedings of the IEEE/CVF Conference on Computer Vision and Pattern Recognition}}. \bibinfo{pages}{4370--4379}.
\newblock


\bibitem[He et~al\mbox{.}(2020)]%
        {he2020momentum}
\bibfield{author}{\bibinfo{person}{Kaiming He}, \bibinfo{person}{Haoqi Fan}, \bibinfo{person}{Yuxin Wu}, \bibinfo{person}{Saining Xie}, {and} \bibinfo{person}{Ross Girshick}.} \bibinfo{year}{2020}\natexlab{}.
\newblock \showarticletitle{Momentum contrast for unsupervised visual representation learning}. In \bibinfo{booktitle}{\emph{Proceedings of the IEEE/CVF conference on computer vision and pattern recognition}}. \bibinfo{pages}{9729--9738}.
\newblock


\bibitem[He et~al\mbox{.}(2016)]%
        {he2016deep}
\bibfield{author}{\bibinfo{person}{Kaiming He}, \bibinfo{person}{Xiangyu Zhang}, \bibinfo{person}{Shaoqing Ren}, {and} \bibinfo{person}{Jian Sun}.} \bibinfo{year}{2016}\natexlab{}.
\newblock \showarticletitle{Deep residual learning for image recognition}. In \bibinfo{booktitle}{\emph{Proceedings of the IEEE conference on computer vision and pattern recognition}}. \bibinfo{pages}{770--778}.
\newblock


\bibitem[He et~al\mbox{.}(2019)]%
        {he2019geonet}
\bibfield{author}{\bibinfo{person}{Tong He}, \bibinfo{person}{Haibin Huang}, \bibinfo{person}{Li Yi}, \bibinfo{person}{Yuqian Zhou}, \bibinfo{person}{Chihao Wu}, \bibinfo{person}{Jue Wang}, {and} \bibinfo{person}{Stefano Soatto}.} \bibinfo{year}{2019}\natexlab{}.
\newblock \showarticletitle{Geonet: Deep geodesic networks for point cloud analysis}. In \bibinfo{booktitle}{\emph{Proceedings of the IEEE/CVF Conference on Computer Vision and Pattern Recognition}}. \bibinfo{pages}{6888--6897}.
\newblock


\bibitem[Jiang et~al\mbox{.}(2023)]%
        {jiang2023neural}
\bibfield{author}{\bibinfo{person}{Puhua Jiang}, \bibinfo{person}{Mingze Sun}, {and} \bibinfo{person}{Ruqi Huang}.} \bibinfo{year}{2023}\natexlab{}.
\newblock \showarticletitle{Neural intrinsic embedding for non-rigid point cloud matching}. In \bibinfo{booktitle}{\emph{Proceedings of the IEEE/CVF Conference on Computer Vision and Pattern Recognition}}. \bibinfo{pages}{21835--21845}.
\newblock


\bibitem[Kanazawa et~al\mbox{.}(2018)]%
        {kanazawa2018end}
\bibfield{author}{\bibinfo{person}{Angjoo Kanazawa}, \bibinfo{person}{Michael~J Black}, \bibinfo{person}{David~W Jacobs}, {and} \bibinfo{person}{Jitendra Malik}.} \bibinfo{year}{2018}\natexlab{}.
\newblock \showarticletitle{End-to-end recovery of human shape and pose}. In \bibinfo{booktitle}{\emph{Proceedings of the IEEE conference on computer vision and pattern recognition}}. \bibinfo{pages}{7122--7131}.
\newblock


\bibitem[Kerbl et~al\mbox{.}(2023)]%
        {kerbl20233d}
\bibfield{author}{\bibinfo{person}{Bernhard Kerbl}, \bibinfo{person}{Georgios Kopanas}, \bibinfo{person}{Thomas Leimk{\"u}hler}, {and} \bibinfo{person}{George Drettakis}.} \bibinfo{year}{2023}\natexlab{}.
\newblock \showarticletitle{3d gaussian splatting for real-time radiance field rendering.}
\newblock \bibinfo{journal}{\emph{ACM Trans. Graph.}} \bibinfo{volume}{42}, \bibinfo{number}{4} (\bibinfo{year}{2023}), \bibinfo{pages}{139--1}.
\newblock


\bibitem[Lang et~al\mbox{.}(2021)]%
        {lang2021dpc}
\bibfield{author}{\bibinfo{person}{Itai Lang}, \bibinfo{person}{Dvir Ginzburg}, \bibinfo{person}{Shai Avidan}, {and} \bibinfo{person}{Dan Raviv}.} \bibinfo{year}{2021}\natexlab{}.
\newblock \showarticletitle{Dpc: Unsupervised deep point correspondence via cross and self construction}. In \bibinfo{booktitle}{\emph{2021 International Conference on 3D Vision (3DV)}}. IEEE, \bibinfo{pages}{1442--1451}.
\newblock


\bibitem[Li et~al\mbox{.}(2021)]%
        {li20214dcomplete}
\bibfield{author}{\bibinfo{person}{Yang Li}, \bibinfo{person}{Hikari Takehara}, \bibinfo{person}{Takafumi Taketomi}, \bibinfo{person}{Bo Zheng}, {and} \bibinfo{person}{Matthias Nie{\ss}ner}.} \bibinfo{year}{2021}\natexlab{}.
\newblock \showarticletitle{4dcomplete: Non-rigid motion estimation beyond the observable surface}. In \bibinfo{booktitle}{\emph{Proceedings of the IEEE/CVF International Conference on Computer Vision}}. \bibinfo{pages}{12706--12716}.
\newblock


\bibitem[Litany et~al\mbox{.}(2017)]%
        {litany2017deep}
\bibfield{author}{\bibinfo{person}{Or Litany}, \bibinfo{person}{Tal Remez}, \bibinfo{person}{Emanuele Rodola}, \bibinfo{person}{Alex Bronstein}, {and} \bibinfo{person}{Michael Bronstein}.} \bibinfo{year}{2017}\natexlab{}.
\newblock \showarticletitle{Deep functional maps: Structured prediction for dense shape correspondence}. In \bibinfo{booktitle}{\emph{Proceedings of the IEEE international conference on computer vision}}. \bibinfo{pages}{5659--5667}.
\newblock


\bibitem[Liu et~al\mbox{.}(2025)]%
        {liu2025stable}
\bibfield{author}{\bibinfo{person}{Haolin Liu}, \bibinfo{person}{Xiaohang Zhan}, \bibinfo{person}{Zizheng Yan}, \bibinfo{person}{Zhongjin Luo}, \bibinfo{person}{Yuxin Wen}, {and} \bibinfo{person}{Xiaoguang Han}.} \bibinfo{year}{2025}\natexlab{}.
\newblock \showarticletitle{Stable-SCore: A Stable Registration-based Framework for 3D Shape Correspondence}.
\newblock \bibinfo{journal}{\emph{arXiv preprint arXiv:2503.21766}} (\bibinfo{year}{2025}).
\newblock


\bibitem[Liu et~al\mbox{.}(2023)]%
        {liu2023zero}
\bibfield{author}{\bibinfo{person}{Ruoshi Liu}, \bibinfo{person}{Rundi Wu}, \bibinfo{person}{Basile Van~Hoorick}, \bibinfo{person}{Pavel Tokmakov}, \bibinfo{person}{Sergey Zakharov}, {and} \bibinfo{person}{Carl Vondrick}.} \bibinfo{year}{2023}\natexlab{}.
\newblock \showarticletitle{Zero-1-to-3: Zero-shot one image to 3d object}. In \bibinfo{booktitle}{\emph{Proceedings of the IEEE/CVF international conference on computer vision}}. \bibinfo{pages}{9298--9309}.
\newblock


\bibitem[Loper et~al\mbox{.}(2023)]%
        {loper2023smpl}
\bibfield{author}{\bibinfo{person}{Matthew Loper}, \bibinfo{person}{Naureen Mahmood}, \bibinfo{person}{Javier Romero}, \bibinfo{person}{Gerard Pons-Moll}, {and} \bibinfo{person}{Michael~J Black}.} \bibinfo{year}{2023}\natexlab{}.
\newblock \showarticletitle{SMPL: A skinned multi-person linear model}.
\newblock In \bibinfo{booktitle}{\emph{Seminal Graphics Papers: Pushing the Boundaries, Volume 2}}. \bibinfo{pages}{851--866}.
\newblock


\bibitem[Loshchilov and Hutter(2017)]%
        {loshchilov2017decoupled}
\bibfield{author}{\bibinfo{person}{Ilya Loshchilov} {and} \bibinfo{person}{Frank Hutter}.} \bibinfo{year}{2017}\natexlab{}.
\newblock \showarticletitle{Decoupled weight decay regularization}.
\newblock \bibinfo{journal}{\emph{arXiv preprint arXiv:1711.05101}} (\bibinfo{year}{2017}).
\newblock


\bibitem[Lu et~al\mbox{.}(2019)]%
        {lu2019deepvcp}
\bibfield{author}{\bibinfo{person}{Weixin Lu}, \bibinfo{person}{Guowei Wan}, \bibinfo{person}{Yao Zhou}, \bibinfo{person}{Xiangyu Fu}, \bibinfo{person}{Pengfei Yuan}, {and} \bibinfo{person}{Shiyu Song}.} \bibinfo{year}{2019}\natexlab{}.
\newblock \showarticletitle{Deepvcp: An end-to-end deep neural network for point cloud registration}. In \bibinfo{booktitle}{\emph{Proceedings of the IEEE/CVF international conference on computer vision}}. \bibinfo{pages}{12--21}.
\newblock


\bibitem[Luo et~al\mbox{.}(2023)]%
        {luo2023diffusion}
\bibfield{author}{\bibinfo{person}{Grace Luo}, \bibinfo{person}{Lisa Dunlap}, \bibinfo{person}{Dong~Huk Park}, \bibinfo{person}{Aleksander Holynski}, {and} \bibinfo{person}{Trevor Darrell}.} \bibinfo{year}{2023}\natexlab{}.
\newblock \showarticletitle{Diffusion hyperfeatures: Searching through time and space for semantic correspondence}.
\newblock \bibinfo{journal}{\emph{Advances in Neural Information Processing Systems}}  \bibinfo{volume}{36} (\bibinfo{year}{2023}), \bibinfo{pages}{47500--47510}.
\newblock


\bibitem[Mahmood et~al\mbox{.}(2019)]%
        {mahmood2019amass}
\bibfield{author}{\bibinfo{person}{Naureen Mahmood}, \bibinfo{person}{Nima Ghorbani}, \bibinfo{person}{Nikolaus~F Troje}, \bibinfo{person}{Gerard Pons-Moll}, {and} \bibinfo{person}{Michael~J Black}.} \bibinfo{year}{2019}\natexlab{}.
\newblock \showarticletitle{AMASS: Archive of motion capture as surface shapes}. In \bibinfo{booktitle}{\emph{Proceedings of the IEEE/CVF international conference on computer vision}}. \bibinfo{pages}{5442--5451}.
\newblock


\bibitem[Mariotti et~al\mbox{.}(2024)]%
        {mariotti2024improving}
\bibfield{author}{\bibinfo{person}{Octave Mariotti}, \bibinfo{person}{Oisin Mac~Aodha}, {and} \bibinfo{person}{Hakan Bilen}.} \bibinfo{year}{2024}\natexlab{}.
\newblock \showarticletitle{Improving semantic correspondence with viewpoint-guided spherical maps}. In \bibinfo{booktitle}{\emph{Proceedings of the IEEE/CVF Conference on Computer Vision and Pattern Recognition}}. \bibinfo{pages}{19521--19530}.
\newblock


\bibitem[Melzi et~al\mbox{.}(2019)]%
        {melzi2019shrec}
\bibfield{author}{\bibinfo{person}{Simone Melzi}, \bibinfo{person}{Riccardo Marin}, \bibinfo{person}{Emanuele Rodol{\`a}}, \bibinfo{person}{Umberto Castellani}, \bibinfo{person}{Jing Ren}, \bibinfo{person}{Adrien Poulenard}, \bibinfo{person}{P Ovsjanikov}, {et~al\mbox{.}}} \bibinfo{year}{2019}\natexlab{}.
\newblock \showarticletitle{SHREC’19: matching humans with different connectivity}. In \bibinfo{booktitle}{\emph{Eurographics Workshop on 3D Object Retrieval}}. The Eurographics Association, \bibinfo{pages}{1--8}.
\newblock


\bibitem[M{\'e}moli and Sapiro(2005)]%
        {memoli2005theoretical}
\bibfield{author}{\bibinfo{person}{Facundo M{\'e}moli} {and} \bibinfo{person}{Guillermo Sapiro}.} \bibinfo{year}{2005}\natexlab{}.
\newblock \showarticletitle{A theoretical and computational framework for isometry invariant recognition of point cloud data}.
\newblock \bibinfo{journal}{\emph{Foundations of Computational Mathematics}}  \bibinfo{volume}{5} (\bibinfo{year}{2005}), \bibinfo{pages}{313--347}.
\newblock


\bibitem[Morreale et~al\mbox{.}(2024)]%
        {morreale2024neural}
\bibfield{author}{\bibinfo{person}{Luca Morreale}, \bibinfo{person}{Noam Aigerman}, \bibinfo{person}{Vladimir~G Kim}, {and} \bibinfo{person}{Niloy~J Mitra}.} \bibinfo{year}{2024}\natexlab{}.
\newblock \showarticletitle{Neural semantic surface maps}. In \bibinfo{booktitle}{\emph{Computer Graphics Forum}}, Vol.~\bibinfo{volume}{43}. Wiley Online Library, \bibinfo{pages}{e15005}.
\newblock


\bibitem[Muralikrishnan et~al\mbox{.}(2024)]%
        {muralikrishnan2024temporal}
\bibfield{author}{\bibinfo{person}{Sanjeev Muralikrishnan}, \bibinfo{person}{Niladri Dutt}, \bibinfo{person}{Siddhartha Chaudhuri}, \bibinfo{person}{Noam Aigerman}, \bibinfo{person}{Vladimir Kim}, \bibinfo{person}{Matthew Fisher}, {and} \bibinfo{person}{Niloy~J Mitra}.} \bibinfo{year}{2024}\natexlab{}.
\newblock \showarticletitle{Temporal Residual Jacobians for Rig-Free Motion Transfer}. In \bibinfo{booktitle}{\emph{European Conference on Computer Vision}}. Springer, \bibinfo{pages}{93--109}.
\newblock


\bibitem[Oord et~al\mbox{.}(2018)]%
        {oord2018representation}
\bibfield{author}{\bibinfo{person}{Aaron van~den Oord}, \bibinfo{person}{Yazhe Li}, {and} \bibinfo{person}{Oriol Vinyals}.} \bibinfo{year}{2018}\natexlab{}.
\newblock \showarticletitle{Representation learning with contrastive predictive coding}.
\newblock \bibinfo{journal}{\emph{arXiv preprint arXiv:1807.03748}} (\bibinfo{year}{2018}).
\newblock


\bibitem[Oquab et~al\mbox{.}(2024)]%
        {oquab2024dinov}
\bibfield{author}{\bibinfo{person}{Maxime Oquab}, \bibinfo{person}{Timoth{\'e}e Darcet}, \bibinfo{person}{Th{\'e}o Moutakanni}, \bibinfo{person}{Huy~V. Vo}, \bibinfo{person}{Marc Szafraniec}, \bibinfo{person}{Vasil Khalidov}, \bibinfo{person}{Pierre Fernandez}, \bibinfo{person}{Daniel HAZIZA}, \bibinfo{person}{Francisco Massa}, \bibinfo{person}{Alaaeldin El-Nouby}, \bibinfo{person}{Mido Assran}, \bibinfo{person}{Nicolas Ballas}, \bibinfo{person}{Wojciech Galuba}, \bibinfo{person}{Russell Howes}, \bibinfo{person}{Po-Yao Huang}, \bibinfo{person}{Shang-Wen Li}, \bibinfo{person}{Ishan Misra}, \bibinfo{person}{Michael Rabbat}, \bibinfo{person}{Vasu Sharma}, \bibinfo{person}{Gabriel Synnaeve}, \bibinfo{person}{Hu Xu}, \bibinfo{person}{Herve Jegou}, \bibinfo{person}{Julien Mairal}, \bibinfo{person}{Patrick Labatut}, \bibinfo{person}{Armand Joulin}, {and} \bibinfo{person}{Piotr Bojanowski}.} \bibinfo{year}{2024}\natexlab{}.
\newblock \showarticletitle{{DINO}v2: Learning Robust Visual Features without Supervision}.
\newblock \bibinfo{journal}{\emph{Transactions on Machine Learning Research}} (\bibinfo{year}{2024}).
\newblock
\showISSN{2835-8856}
\urldef\tempurl%
\url{https://openreview.net/forum?id=a68SUt6zFt}
\showURL{%
\tempurl}
\newblock
\shownote{Featured Certification}.


\bibitem[Ovsjanikov et~al\mbox{.}(2012)]%
        {ovsjanikov2012functional}
\bibfield{author}{\bibinfo{person}{Maks Ovsjanikov}, \bibinfo{person}{Mirela Ben-Chen}, \bibinfo{person}{Justin Solomon}, \bibinfo{person}{Adrian Butscher}, {and} \bibinfo{person}{Leonidas Guibas}.} \bibinfo{year}{2012}\natexlab{}.
\newblock \showarticletitle{Functional maps: a flexible representation of maps between shapes}.
\newblock \bibinfo{journal}{\emph{ACM Transactions on Graphics (ToG)}} \bibinfo{volume}{31}, \bibinfo{number}{4} (\bibinfo{year}{2012}), \bibinfo{pages}{1--11}.
\newblock


\bibitem[Peng et~al\mbox{.}(2019)]%
        {peng2019pvnet}
\bibfield{author}{\bibinfo{person}{Sida Peng}, \bibinfo{person}{Yuan Liu}, \bibinfo{person}{Qixing Huang}, \bibinfo{person}{Xiaowei Zhou}, {and} \bibinfo{person}{Hujun Bao}.} \bibinfo{year}{2019}\natexlab{}.
\newblock \showarticletitle{Pvnet: Pixel-wise voting network for 6dof pose estimation}. In \bibinfo{booktitle}{\emph{Proceedings of the IEEE/CVF conference on computer vision and pattern recognition}}. \bibinfo{pages}{4561--4570}.
\newblock


\bibitem[Polyak and Juditsky(1992)]%
        {polyak1992acceleration}
\bibfield{author}{\bibinfo{person}{Boris~T Polyak} {and} \bibinfo{person}{Anatoli~B Juditsky}.} \bibinfo{year}{1992}\natexlab{}.
\newblock \showarticletitle{Acceleration of stochastic approximation by averaging}.
\newblock \bibinfo{journal}{\emph{SIAM journal on control and optimization}} \bibinfo{volume}{30}, \bibinfo{number}{4} (\bibinfo{year}{1992}), \bibinfo{pages}{838--855}.
\newblock


\bibitem[Poole et~al\mbox{.}(2022)]%
        {poole2022dreamfusion}
\bibfield{author}{\bibinfo{person}{Ben Poole}, \bibinfo{person}{Ajay Jain}, \bibinfo{person}{Jonathan~T Barron}, {and} \bibinfo{person}{Ben Mildenhall}.} \bibinfo{year}{2022}\natexlab{}.
\newblock \showarticletitle{Dreamfusion: Text-to-3d using 2d diffusion}.
\newblock \bibinfo{journal}{\emph{arXiv preprint arXiv:2209.14988}} (\bibinfo{year}{2022}).
\newblock


\bibitem[Radford et~al\mbox{.}(2021)]%
        {radford2021learning}
\bibfield{author}{\bibinfo{person}{Alec Radford}, \bibinfo{person}{Jong~Wook Kim}, \bibinfo{person}{Chris Hallacy}, \bibinfo{person}{Aditya Ramesh}, \bibinfo{person}{Gabriel Goh}, \bibinfo{person}{Sandhini Agarwal}, \bibinfo{person}{Girish Sastry}, \bibinfo{person}{Amanda Askell}, \bibinfo{person}{Pamela Mishkin}, \bibinfo{person}{Jack Clark}, {et~al\mbox{.}}} \bibinfo{year}{2021}\natexlab{}.
\newblock \showarticletitle{Learning transferable visual models from natural language supervision}. In \bibinfo{booktitle}{\emph{International conference on machine learning}}. PmLR, \bibinfo{pages}{8748--8763}.
\newblock


\bibitem[Rodol{\`a} et~al\mbox{.}(2015)]%
        {rodola2015point}
\bibfield{author}{\bibinfo{person}{Emanuele Rodol{\`a}}, \bibinfo{person}{Michael Moeller}, {and} \bibinfo{person}{Daniel Cremers}.} \bibinfo{year}{2015}\natexlab{}.
\newblock \showarticletitle{Point-wise map recovery and refinement from functional correspondence}.
\newblock \bibinfo{journal}{\emph{arXiv preprint arXiv:1506.05603}} (\bibinfo{year}{2015}).
\newblock


\bibitem[Rombach et~al\mbox{.}(2022)]%
        {rombach2022high}
\bibfield{author}{\bibinfo{person}{Robin Rombach}, \bibinfo{person}{Andreas Blattmann}, \bibinfo{person}{Dominik Lorenz}, \bibinfo{person}{Patrick Esser}, {and} \bibinfo{person}{Bj{\"o}rn Ommer}.} \bibinfo{year}{2022}\natexlab{}.
\newblock \showarticletitle{High-resolution image synthesis with latent diffusion models}. In \bibinfo{booktitle}{\emph{Proceedings of the IEEE/CVF conference on computer vision and pattern recognition}}. \bibinfo{pages}{10684--10695}.
\newblock


\bibitem[Roufosse et~al\mbox{.}(2019)]%
        {roufosse2019unsupervised}
\bibfield{author}{\bibinfo{person}{Jean-Michel Roufosse}, \bibinfo{person}{Abhishek Sharma}, {and} \bibinfo{person}{Maks Ovsjanikov}.} \bibinfo{year}{2019}\natexlab{}.
\newblock \showarticletitle{Unsupervised deep learning for structured shape matching}. In \bibinfo{booktitle}{\emph{Proceedings of the IEEE/CVF International Conference on Computer Vision}}. \bibinfo{pages}{1617--1627}.
\newblock


\bibitem[Schroff et~al\mbox{.}(2015)]%
        {schroff2015facenet}
\bibfield{author}{\bibinfo{person}{Florian Schroff}, \bibinfo{person}{Dmitry Kalenichenko}, {and} \bibinfo{person}{James Philbin}.} \bibinfo{year}{2015}\natexlab{}.
\newblock \showarticletitle{Facenet: A unified embedding for face recognition and clustering}. In \bibinfo{booktitle}{\emph{Proceedings of the IEEE conference on computer vision and pattern recognition}}. \bibinfo{pages}{815--823}.
\newblock


\bibitem[Shamai and Kimmel(2017)]%
        {shamai2017geodesic}
\bibfield{author}{\bibinfo{person}{Gil Shamai} {and} \bibinfo{person}{Ron Kimmel}.} \bibinfo{year}{2017}\natexlab{}.
\newblock \showarticletitle{Geodesic distance descriptors}. In \bibinfo{booktitle}{\emph{Proceedings of the IEEE Conference on Computer Vision and Pattern Recognition}}. \bibinfo{pages}{6410--6418}.
\newblock


\bibitem[Sharp et~al\mbox{.}(2019b)]%
        {polyscope}
\bibfield{author}{\bibinfo{person}{Nicholas Sharp} {et~al\mbox{.}}} \bibinfo{year}{2019}\natexlab{b}.
\newblock \bibinfo{title}{Polyscope}.
\newblock
\newblock
\shownote{www.polyscope.run}.


\bibitem[Sharp et~al\mbox{.}(2019a)]%
        {geometrycentral}
\bibfield{author}{\bibinfo{person}{Nicholas Sharp}, \bibinfo{person}{Keenan Crane}, {et~al\mbox{.}}} \bibinfo{year}{2019}\natexlab{a}.
\newblock \showarticletitle{GeometryCentral: A modern C++ library of data structures and algorithms for geometry processing}.
\newblock \bibinfo{howpublished}{\url{https://geometry-central.net/}}.
\newblock  (\bibinfo{year}{2019}).
\newblock


\bibitem[Shtedritski et~al\mbox{.}(2024)]%
        {shtedritski2024shic}
\bibfield{author}{\bibinfo{person}{Aleksandar Shtedritski}, \bibinfo{person}{Christian Rupprecht}, {and} \bibinfo{person}{Andrea Vedaldi}.} \bibinfo{year}{2024}\natexlab{}.
\newblock \showarticletitle{SHIC: Shape-Image Correspondences with No Keypoint Supervision}. In \bibinfo{booktitle}{\emph{European Conference on Computer Vision}}. Springer, \bibinfo{pages}{129--145}.
\newblock


\bibitem[Sorkine and Alexa(2007)]%
        {sorkine2007rigid}
\bibfield{author}{\bibinfo{person}{Olga Sorkine} {and} \bibinfo{person}{Marc Alexa}.} \bibinfo{year}{2007}\natexlab{}.
\newblock \showarticletitle{As-rigid-as-possible surface modeling}. In \bibinfo{booktitle}{\emph{Symposium on Geometry processing}}, Vol.~\bibinfo{volume}{4}. Citeseer, \bibinfo{pages}{109--116}.
\newblock


\bibitem[Sun et~al\mbox{.}(2023a)]%
        {sun2023mapconnet}
\bibfield{author}{\bibinfo{person}{Jiaze Sun}, \bibinfo{person}{Zhixiang Chen}, {and} \bibinfo{person}{Tae-Kyun Kim}.} \bibinfo{year}{2023}\natexlab{a}.
\newblock \showarticletitle{Mapconnet: Self-supervised 3d pose transfer with mesh and point contrastive learning}. In \bibinfo{booktitle}{\emph{Proceedings of the IEEE/CVF International Conference on Computer Vision}}. \bibinfo{pages}{14452--14462}.
\newblock


\bibitem[Sun et~al\mbox{.}(2009)]%
        {sun2009concise}
\bibfield{author}{\bibinfo{person}{Jian Sun}, \bibinfo{person}{Maks Ovsjanikov}, {and} \bibinfo{person}{Leonidas Guibas}.} \bibinfo{year}{2009}\natexlab{}.
\newblock \showarticletitle{A concise and provably informative multi-scale signature based on heat diffusion}. In \bibinfo{booktitle}{\emph{Computer graphics forum}}, Vol.~\bibinfo{volume}{28}. Wiley Online Library, \bibinfo{pages}{1383--1392}.
\newblock


\bibitem[Sun et~al\mbox{.}(2024)]%
        {sun2024srif}
\bibfield{author}{\bibinfo{person}{Mingze Sun}, \bibinfo{person}{Chen Guo}, \bibinfo{person}{Puhua Jiang}, \bibinfo{person}{Shiwei Mao}, \bibinfo{person}{Yurun Chen}, {and} \bibinfo{person}{Ruqi Huang}.} \bibinfo{year}{2024}\natexlab{}.
\newblock \showarticletitle{SRIF: Semantic Shape Registration Empowered by Diffusion-based Image Morphing and Flow Estimation}. In \bibinfo{booktitle}{\emph{SIGGRAPH Asia 2024 Conference Papers}}. \bibinfo{pages}{1--11}.
\newblock


\bibitem[Sun et~al\mbox{.}(2023b)]%
        {sun2023spatially}
\bibfield{author}{\bibinfo{person}{Mingze Sun}, \bibinfo{person}{Shiwei Mao}, \bibinfo{person}{Puhua Jiang}, \bibinfo{person}{Maks Ovsjanikov}, {and} \bibinfo{person}{Ruqi Huang}.} \bibinfo{year}{2023}\natexlab{b}.
\newblock \showarticletitle{Spatially and spectrally consistent deep functional maps}. In \bibinfo{booktitle}{\emph{Proceedings of the IEEE/CVF International Conference on Computer Vision}}. \bibinfo{pages}{14497--14507}.
\newblock


\bibitem[Tang et~al\mbox{.}(2023)]%
        {tang2023emergent}
\bibfield{author}{\bibinfo{person}{Luming Tang}, \bibinfo{person}{Menglin Jia}, \bibinfo{person}{Qianqian Wang}, \bibinfo{person}{Cheng~Perng Phoo}, {and} \bibinfo{person}{Bharath Hariharan}.} \bibinfo{year}{2023}\natexlab{}.
\newblock \showarticletitle{Emergent correspondence from image diffusion}.
\newblock \bibinfo{journal}{\emph{Advances in Neural Information Processing Systems}}  \bibinfo{volume}{36} (\bibinfo{year}{2023}), \bibinfo{pages}{1363--1389}.
\newblock


\bibitem[Tombari et~al\mbox{.}(2010)]%
        {tombari2010unique}
\bibfield{author}{\bibinfo{person}{Federico Tombari}, \bibinfo{person}{Samuele Salti}, {and} \bibinfo{person}{Luigi Di~Stefano}.} \bibinfo{year}{2010}\natexlab{}.
\newblock \showarticletitle{Unique signatures of histograms for local surface description}. In \bibinfo{booktitle}{\emph{Computer Vision--ECCV 2010: 11th European Conference on Computer Vision, Heraklion, Crete, Greece, September 5-11, 2010, Proceedings, Part III 11}}. Springer, \bibinfo{pages}{356--369}.
\newblock


\bibitem[Uzolas et~al\mbox{.}(2023)]%
        {uzolas2023template}
\bibfield{author}{\bibinfo{person}{Lukas Uzolas}, \bibinfo{person}{Elmar Eisemann}, {and} \bibinfo{person}{Petr Kellnhofer}.} \bibinfo{year}{2023}\natexlab{}.
\newblock \showarticletitle{Template-free articulated neural point clouds for reposable view synthesis}.
\newblock \bibinfo{journal}{\emph{Advances in Neural Information Processing Systems}}  \bibinfo{volume}{36} (\bibinfo{year}{2023}), \bibinfo{pages}{31621--31637}.
\newblock


\bibitem[Uzolas et~al\mbox{.}(2024)]%
        {uzolas2024motiondreamer}
\bibfield{author}{\bibinfo{person}{Lukas Uzolas}, \bibinfo{person}{Elmar Eisemann}, {and} \bibinfo{person}{Petr Kellnhofer}.} \bibinfo{year}{2024}\natexlab{}.
\newblock \showarticletitle{Motiondreamer: Zero-shot 3d mesh animation from video diffusion models}.
\newblock \bibinfo{journal}{\emph{arXiv preprint arXiv:2405.20155}} (\bibinfo{year}{2024}).
\newblock


\bibitem[Wang et~al\mbox{.}(2022)]%
        {wang2022dual}
\bibfield{author}{\bibinfo{person}{Peng-Shuai Wang}, \bibinfo{person}{Yang Liu}, {and} \bibinfo{person}{Xin Tong}.} \bibinfo{year}{2022}\natexlab{}.
\newblock \showarticletitle{Dual octree graph networks for learning adaptive volumetric shape representations}.
\newblock \bibinfo{journal}{\emph{ACM Transactions on Graphics (TOG)}} \bibinfo{volume}{41}, \bibinfo{number}{4} (\bibinfo{year}{2022}), \bibinfo{pages}{1--15}.
\newblock


\bibitem[Wang and Solomon(2019)]%
        {wang2019deep}
\bibfield{author}{\bibinfo{person}{Yue Wang} {and} \bibinfo{person}{Justin~M Solomon}.} \bibinfo{year}{2019}\natexlab{}.
\newblock \showarticletitle{Deep closest point: Learning representations for point cloud registration}. In \bibinfo{booktitle}{\emph{Proceedings of the IEEE/CVF international conference on computer vision}}. \bibinfo{pages}{3523--3532}.
\newblock


\bibitem[Weinberger and Saul(2009)]%
        {weinberger2009distance}
\bibfield{author}{\bibinfo{person}{Kilian~Q Weinberger} {and} \bibinfo{person}{Lawrence~K Saul}.} \bibinfo{year}{2009}\natexlab{}.
\newblock \showarticletitle{Distance metric learning for large margin nearest neighbor classification.}
\newblock \bibinfo{journal}{\emph{Journal of machine learning research}} \bibinfo{volume}{10}, \bibinfo{number}{2} (\bibinfo{year}{2009}).
\newblock


\bibitem[Wimmer et~al\mbox{.}(2024)]%
        {wimmer2024back}
\bibfield{author}{\bibinfo{person}{Thomas Wimmer}, \bibinfo{person}{Peter Wonka}, {and} \bibinfo{person}{Maks Ovsjanikov}.} \bibinfo{year}{2024}\natexlab{}.
\newblock \showarticletitle{Back to 3D: Few-Shot 3D Keypoint Detection with Back-Projected 2D Features}. In \bibinfo{booktitle}{\emph{Proceedings of the IEEE/CVF Conference on Computer Vision and Pattern Recognition}}. \bibinfo{pages}{4154--4164}.
\newblock


\bibitem[Wu et~al\mbox{.}(2018)]%
        {wu2018unsupervised}
\bibfield{author}{\bibinfo{person}{Zhirong Wu}, \bibinfo{person}{Yuanjun Xiong}, \bibinfo{person}{Stella~X Yu}, {and} \bibinfo{person}{Dahua Lin}.} \bibinfo{year}{2018}\natexlab{}.
\newblock \showarticletitle{Unsupervised feature learning via non-parametric instance discrimination}. In \bibinfo{booktitle}{\emph{Proceedings of the IEEE conference on computer vision and pattern recognition}}. \bibinfo{pages}{3733--3742}.
\newblock


\bibitem[Xia et~al\mbox{.}(2021)]%
        {xia2021geodesicembedding}
\bibfield{author}{\bibinfo{person}{Qianwei Xia}, \bibinfo{person}{Juyong Zhang}, \bibinfo{person}{Zheng Fang}, \bibinfo{person}{Jin Li}, \bibinfo{person}{Mingyue Zhang}, \bibinfo{person}{Bailin Deng}, {and} \bibinfo{person}{Ying He}.} \bibinfo{year}{2021}\natexlab{}.
\newblock \showarticletitle{GeodesicEmbedding (GE): a high-dimensional embedding approach for fast geodesic distance queries}.
\newblock \bibinfo{journal}{\emph{IEEE Transactions on Visualization and Computer Graphics}} \bibinfo{volume}{28}, \bibinfo{number}{12} (\bibinfo{year}{2021}), \bibinfo{pages}{4930--4939}.
\newblock


\bibitem[Xu et~al\mbox{.}(2020)]%
        {xu2020rignet}
\bibfield{author}{\bibinfo{person}{Zhan Xu}, \bibinfo{person}{Yang Zhou}, \bibinfo{person}{Evangelos Kalogerakis}, \bibinfo{person}{Chris Landreth}, {and} \bibinfo{person}{Karan Singh}.} \bibinfo{year}{2020}\natexlab{}.
\newblock \showarticletitle{Rignet: Neural rigging for articulated characters}.
\newblock \bibinfo{journal}{\emph{arXiv preprint arXiv:2005.00559}} (\bibinfo{year}{2020}).
\newblock


\bibitem[Yang et~al\mbox{.}(2015)]%
        {yang2015go}
\bibfield{author}{\bibinfo{person}{Jiaolong Yang}, \bibinfo{person}{Hongdong Li}, \bibinfo{person}{Dylan Campbell}, {and} \bibinfo{person}{Yunde Jia}.} \bibinfo{year}{2015}\natexlab{}.
\newblock \showarticletitle{Go-ICP: A globally optimal solution to 3D ICP point-set registration}.
\newblock \bibinfo{journal}{\emph{IEEE transactions on pattern analysis and machine intelligence}} \bibinfo{volume}{38}, \bibinfo{number}{11} (\bibinfo{year}{2015}), \bibinfo{pages}{2241--2254}.
\newblock


\bibitem[Yao et~al\mbox{.}(2022)]%
        {yao2022lassie}
\bibfield{author}{\bibinfo{person}{Chun-Han Yao}, \bibinfo{person}{Wei-Chih Hung}, \bibinfo{person}{Yuanzhen Li}, \bibinfo{person}{Michael Rubinstein}, \bibinfo{person}{Ming-Hsuan Yang}, {and} \bibinfo{person}{Varun Jampani}.} \bibinfo{year}{2022}\natexlab{}.
\newblock \showarticletitle{Lassie: Learning articulated shapes from sparse image ensemble via 3d part discovery}.
\newblock \bibinfo{journal}{\emph{Advances in Neural Information Processing Systems}}  \bibinfo{volume}{35} (\bibinfo{year}{2022}), \bibinfo{pages}{15296--15308}.
\newblock


\bibitem[Yew and Lee(2018)]%
        {yew20183dfeat}
\bibfield{author}{\bibinfo{person}{Zi~Jian Yew} {and} \bibinfo{person}{Gim~Hee Lee}.} \bibinfo{year}{2018}\natexlab{}.
\newblock \showarticletitle{3dfeat-net: Weakly supervised local 3d features for point cloud registration}. In \bibinfo{booktitle}{\emph{Proceedings of the European conference on computer vision (ECCV)}}. \bibinfo{pages}{607--623}.
\newblock


\bibitem[Zeng et~al\mbox{.}(2017)]%
        {zeng20173dmatch}
\bibfield{author}{\bibinfo{person}{Andy Zeng}, \bibinfo{person}{Shuran Song}, \bibinfo{person}{Matthias Nie{\ss}ner}, \bibinfo{person}{Matthew Fisher}, \bibinfo{person}{Jianxiong Xiao}, {and} \bibinfo{person}{Thomas Funkhouser}.} \bibinfo{year}{2017}\natexlab{}.
\newblock \showarticletitle{3dmatch: Learning local geometric descriptors from rgb-d reconstructions}. In \bibinfo{booktitle}{\emph{Proceedings of the IEEE conference on computer vision and pattern recognition}}. \bibinfo{pages}{1802--1811}.
\newblock


\bibitem[Zeng et~al\mbox{.}(2021)]%
        {zeng2021corrnet3d}
\bibfield{author}{\bibinfo{person}{Yiming Zeng}, \bibinfo{person}{Yue Qian}, \bibinfo{person}{Zhiyu Zhu}, \bibinfo{person}{Junhui Hou}, \bibinfo{person}{Hui Yuan}, {and} \bibinfo{person}{Ying He}.} \bibinfo{year}{2021}\natexlab{}.
\newblock \showarticletitle{Corrnet3d: Unsupervised end-to-end learning of dense correspondence for 3d point clouds}. In \bibinfo{booktitle}{\emph{Proceedings of the IEEE/CVF Conference on Computer Vision and Pattern Recognition}}. \bibinfo{pages}{6052--6061}.
\newblock


\bibitem[Zhang et~al\mbox{.}(2024a)]%
        {zhang2024telling}
\bibfield{author}{\bibinfo{person}{Junyi Zhang}, \bibinfo{person}{Charles Herrmann}, \bibinfo{person}{Junhwa Hur}, \bibinfo{person}{Eric Chen}, \bibinfo{person}{Varun Jampani}, \bibinfo{person}{Deqing Sun}, {and} \bibinfo{person}{Ming-Hsuan Yang}.} \bibinfo{year}{2024}\natexlab{a}.
\newblock \showarticletitle{Telling left from right: Identifying geometry-aware semantic correspondence}. In \bibinfo{booktitle}{\emph{Proceedings of the IEEE/CVF Conference on Computer Vision and Pattern Recognition}}. \bibinfo{pages}{3076--3085}.
\newblock


\bibitem[Zhang et~al\mbox{.}(2023a)]%
        {zhang2023tale}
\bibfield{author}{\bibinfo{person}{Junyi Zhang}, \bibinfo{person}{Charles Herrmann}, \bibinfo{person}{Junhwa Hur}, \bibinfo{person}{Luisa Polania~Cabrera}, \bibinfo{person}{Varun Jampani}, \bibinfo{person}{Deqing Sun}, {and} \bibinfo{person}{Ming-Hsuan Yang}.} \bibinfo{year}{2023}\natexlab{a}.
\newblock \showarticletitle{A tale of two features: Stable diffusion complements dino for zero-shot semantic correspondence}.
\newblock \bibinfo{journal}{\emph{Advances in Neural Information Processing Systems}}  \bibinfo{volume}{36} (\bibinfo{year}{2023}), \bibinfo{pages}{45533--45547}.
\newblock


\bibitem[Zhang et~al\mbox{.}(2024b)]%
        {zhang2024tale}
\bibfield{author}{\bibinfo{person}{Junyi Zhang}, \bibinfo{person}{Charles Herrmann}, \bibinfo{person}{Junhwa Hur}, \bibinfo{person}{Luisa Polania~Cabrera}, \bibinfo{person}{Varun Jampani}, \bibinfo{person}{Deqing Sun}, {and} \bibinfo{person}{Ming-Hsuan Yang}.} \bibinfo{year}{2024}\natexlab{b}.
\newblock \showarticletitle{A tale of two features: Stable diffusion complements dino for zero-shot semantic correspondence}.
\newblock \bibinfo{journal}{\emph{Advances in Neural Information Processing Systems}}  \bibinfo{volume}{36} (\bibinfo{year}{2024}).
\newblock


\bibitem[Zhang et~al\mbox{.}(2023b)]%
        {zhang2023adding}
\bibfield{author}{\bibinfo{person}{Lvmin Zhang}, \bibinfo{person}{Anyi Rao}, {and} \bibinfo{person}{Maneesh Agrawala}.} \bibinfo{year}{2023}\natexlab{b}.
\newblock \showarticletitle{Adding conditional control to text-to-image diffusion models}. In \bibinfo{booktitle}{\emph{Proceedings of the IEEE/CVF international conference on computer vision}}. \bibinfo{pages}{3836--3847}.
\newblock


\bibitem[Zuffi et~al\mbox{.}(2017)]%
        {zuffi20173d}
\bibfield{author}{\bibinfo{person}{Silvia Zuffi}, \bibinfo{person}{Angjoo Kanazawa}, \bibinfo{person}{David~W Jacobs}, {and} \bibinfo{person}{Michael~J Black}.} \bibinfo{year}{2017}\natexlab{}.
\newblock \showarticletitle{3D menagerie: Modeling the 3D shape and pose of animals}. In \bibinfo{booktitle}{\emph{Proceedings of the IEEE conference on computer vision and pattern recognition}}. \bibinfo{pages}{6365--6373}.
\newblock


\end{thebibliography}


\appendix

\pagenumbering{gobble}
\clearpage

\title{Surface-Aware Distilled 3D Semantic Features: Appendix}

\pagenumbering{gobble}

\makeatletter
\renewcommand{\@teaserfigures}{}
\renewcommand{\@abstract}{}  

\copyrightyear{}
\acmYear{}
\acmConference[]{}{}{}
\acmBooktitle{}\acmDOI{}
\acmISBN{}

\makeatother

\settopmatter{printacmref=false, printccs=false, printfolios=false}

\setcounter{page}{1}
\maketitle

\section{Additional implementation details}
Here, we provide additional details to reproduce our experiments.

\subsection{Diff3F}
\label{supp:diff3f_baselines}
We use the authors' code to compute the Diff3F features~\cite{dutt2024diffusion} for our \emph{base features} and as a baseline method for our comparisons.

Since we observed that the camera poses used for sampling in the \dtosca{} dataset are biased towards a specific up-direction, we modified the code to flip the coordinate system for only this dataset. 
Doing so yields $\approx10$\% increase in correspondence accuracy in \dtosca{} for both our method and the Diff3F baseline, when compared to the numbers reported in the Diff3F paper~\cite{dutt2024diffusion}. 
An alternative solution could be a modification of the camera sampling algorithm itself.

\subsection{Ablations}

\begin{figure}
    \centering
    \includegraphics[width=1\linewidth]{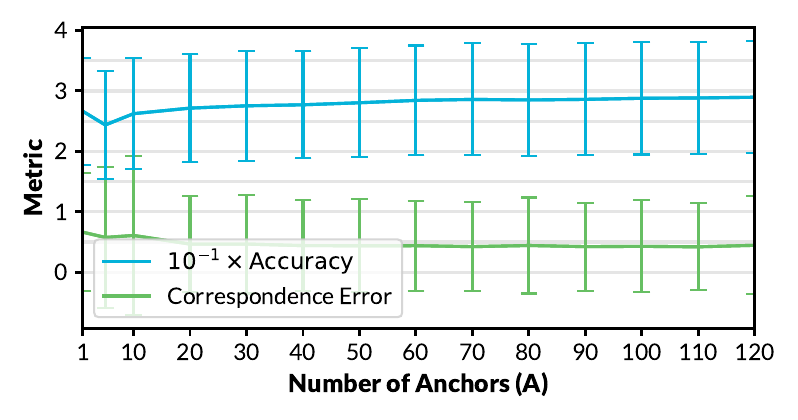}
    \caption{Ablation on the anchor count $A$ on \dshrechumans{}.
    In terms of the correspondence error, our method improves upon Diff3F already starting from $A=1$.
    }
    \label{fig:ablations}
\end{figure}

\paragraph{Number of Anchors} We train our method for different anchor counts $A$ with a constant two-hour training budget per model. 
In \refFig{ablations}, we observe a low sensitivity to the anchor count.
Due to repeated random sampling over the course of training, even $A=1$ outperforms the correspondence error of Diff3F.
Ultimately, we opt for $A=100$ in all our experiments, as it balances computation cost and matching performance well.
Note that neither the parameter $A$ nor any computation of geodesic distances in generally are used during inference.

\paragraph{Losses}

\label{supp:ablations_baselines}
Utilizing the geodesic distance as a supervision signal for embeddings has been explored in related works ~\cite{jiang2023neural,chen2025dvmatcherdeformationbasednonrigidpoint}. However, in our ablations, we show that we achieve superior results with our formulation. In this section, we discuss the key differences.

\paragraph{Relative Geodesic Loss (RGL)}
Based on two points, $v_p$ and $v_q$, the Relative Geodesic Loss~\cite{jiang2023neural} minimizes the difference between the geodesic distance $d^S$ and the \textit{Euclidean} embedding distance $d^E$ of those two points, relative to the surface distance:
\begin{equation}
L_{RGL} = \sum_{i} \sum_{(p,q) \in S_i} \frac{|d_i^E (v_p, v_q) - d^S (v_p, v_q)|^2}{d^S (v_p, v_q)^2}.
\end{equation}
The normalization term is introduced to prioritize local distance preservation.
We do not utilize this normalization term, because the \emph{base features} struggle to disambiguate samples that are far away on the surface but close in feature space, such as ``left hand'' and ``right hand''.

\paragraph{\textit{Naive} Geodesic Loss (NGL)} \textit{NGL} is discussed by Jiang~et~al. \shortcite{jiang2023neural} but not used for training, as the authors state that it might hamper local distance preservation.
Indeed, in our ablations, it achieves worse results in terms of correspondence accuracy when compared to \textit{RGL}.
It is identical to \textit{RGL} but it omits the normalization term:
\begin{equation}
    L_{NGL} = \sum_{i} \sum_{(p,q) \in S_i \in [n_i]^2} |d_i^E (v_p, v_q) - d^S (v_p, v_q)|^2
\end{equation}
While not actually utilized in their work, the \textit{NGL} formulation is the closest of the three to our own formulation. 
However, our choice of a hyperspherical rather than Euclidean embedding space in combination with our autoencoder setup achieves notably better results.

\paragraph{Geometric Similarity Loss (GSL)} 
A concurrent work proposes to maximize the \textit{local angular similarity} between geodesic distances and \emph{Euclidean} feature vectors for a set of neighbors for each point~\cite{chen2025dvmatcherdeformationbasednonrigidpoint}. 
Crucially, the similarity is only maximized for a set of $k$ point pairs nearest in the embedding space. The loss is a cosine metric between a vector of geodesic distances $\mathbf{m}_i \in \reals^k$ and a vector of \emph{Euclidean} embedding distances $\mathbf{d}_i \in \reals^k$:
\begin{equation}
\mathcal{L}_{\text{GSL}} = \frac{1}{N} \sum_{i=1}^{N} \left( 1 - \frac{\mathbf{d}_i \cdot \mathbf{m}_i}{\|\mathbf{d}_i\| \|\mathbf{m}_i\|} \right).
\end{equation}
This restricts \textit{GSL} supervision to a fixed neighborhood size and potentially limits disambiguation of features that are close in feature space but not among the $k$ nearest neighbors. 
In contrast, our method follows a global approach by sampling anchors based on a furthest point sampling.

\paragraph{Conclusion} Our method differs from recent and concurrent works utilizing vision-based features for 3D shape matching in three main aspects: 
1) we follow a global approach when enforcing distances in the embedding space; 
2) our embedding space is hyperspherical and it only encodes angular information; 
3) in the context of the whole pipeline, we solely rely on intrinsic properties.

\section{One Shot Pose Transfer}
\label{supp:pose_transfer}
We train an MLP to model the deformation between the paired input source mesh $\mathcal{M}_{train}^{src}$ and the output target mesh $\mathcal{M}_{train}^{tgt}$ obtained from \damass and thus not used for training of our features. 
The paired training meshes share the same identity $\beta$ but they differ in poses $\theta$ such that
\begin{align*}
\mathcal{M}_{train}^{src} &:= SMPL(\beta_{src}, \theta_{src})\\
\mathcal{M}_{train}^{tgt} &:= SMPL(\beta_{src}, \theta_{tgt}).
\end{align*}

We train an MLP $\mathcal{M}_{train}^{tgt} = \mathcal{J}(\mathcal{M}_{train}^{src}, \mathcal{S_M^{\text{\textit{src}}}})$ to produce the target pose mesh $\mathcal{M}_{train}^{tgt}$ conditioned on the source mesh $\mathcal{M}_{train}^{src}$ and its \emph{surface-aware features} $\mathcal{S_M^{\text{\textit{src}}}}$.
These features are produced by our pre-trained general encoder $\mathcal{E}(.)$ from \refSec{experiments} without any further fine-tuning. Alternatively, other features are used for comparisons.

During training, $\mathcal{J}(.)$ learns a \textit{per-face} residual function ~\cite{muralikrishnan2024temporal} to model a Neural Jacobian Field~\cite{aigerman2022neural} while being supervised by MSE between the predicted re-posed mesh and the target $\mathcal{M}_{train}^{tgt}$.

Crucially, $\mathcal{J}(.)$ is defined on a per-face basis, which means that the input meshes used in test time do not need to have the same connectivity as the training mesh pair.
In quantitative comparisons, we apply the learned mapping $\mathcal{J}(.)$ to input meshes of unseen identities $\mathcal{M}_{test}^{src} := SMPL(\beta_{test}, \theta_{src})$ and compare against their ground truths $\mathcal{M}_{test}^{tgt} := SMPL(\beta_{test}, \theta_{tgt})$. 
For other shape classes, we provide qualitative comparisons.

\section{Pose Alignment}
\label{supp:pose_alignment}
We establish correspondences between two input shapes based on the feature cosine similarity $\phi(.)$ (\refSec{preliminaries}), such that each point $\mathbf{x}_i^\mathcal{S}$ in the source shape is assigned a target point $\mathbf{x}_i^\mathcal{T}$ in the target shape. 
Next, we align the source to the target by minimizing the following L1 loss:
\begin{equation}
    \mathcal{L}_{point} = \frac{1}{N}\sum_i^N||\mathbf{x}_i^\mathcal{S} - \mathbf{x}_i^\mathcal{T}||_1. 
\end{equation}
For the first half of the optimization steps, we only optimize the root rotation $\mathbf{R}$, the root translation $\mathbf{t}$, and the scale $s$, which roughly rigidly aligns the meshes. In the second half, we additionally optimize the rotation $\mathbf{R}_b$ of each kinematic bone $b$. The parameters are optimized based on a gradient-descent for $4000$ iterations, which takes approximately 30 seconds for a static pose.

Furthermore, we found it beneficial to use an as-rigid-as-possible regularization~\cite{sorkine2007rigid}, which penalizes the deviation between the initial edge lengths of the mesh $\delta^{init}_e$ and the current edge length $\delta_e$ for each edge $e$:
\begin{equation}
    \mathcal{L}_{arap} = \frac{1}{E}\sum_e^E|\delta_e^{init} - \delta_e|. 
\end{equation}

When fitting an animation as a pose sequence, we optimize the pose parameters for each time step $t$. Furthermore, we apply pointwise temporal smoothing for neighboring frames:
\begin{equation}
    \mathcal{L}_{smooth} = \frac{1}{N(T-1)}\sum_t^{T-1}\sum_i^N ||\mathbf{x}_{i,t}^\mathcal{S} - \mathbf{x}_{i,t+1}^\mathcal{S} ||_2^2. 
\end{equation}

The final loss is $\mathcal{L}_{pose} = w_p\mathcal{L}_{point} + w_a\mathcal{L}_{arap} + w_s\mathcal{L}_{smooth}$ with $w_p = w_a = w_s = 1$ for animations and $w_s=0$ otherwise.

\section{Additional results}

\subsection{Qualitative Results on SHREC'20}
\label{supp:qualitative_shrec20}
\begin{figure*}
    \centering
    \includegraphics[width=\textwidth]{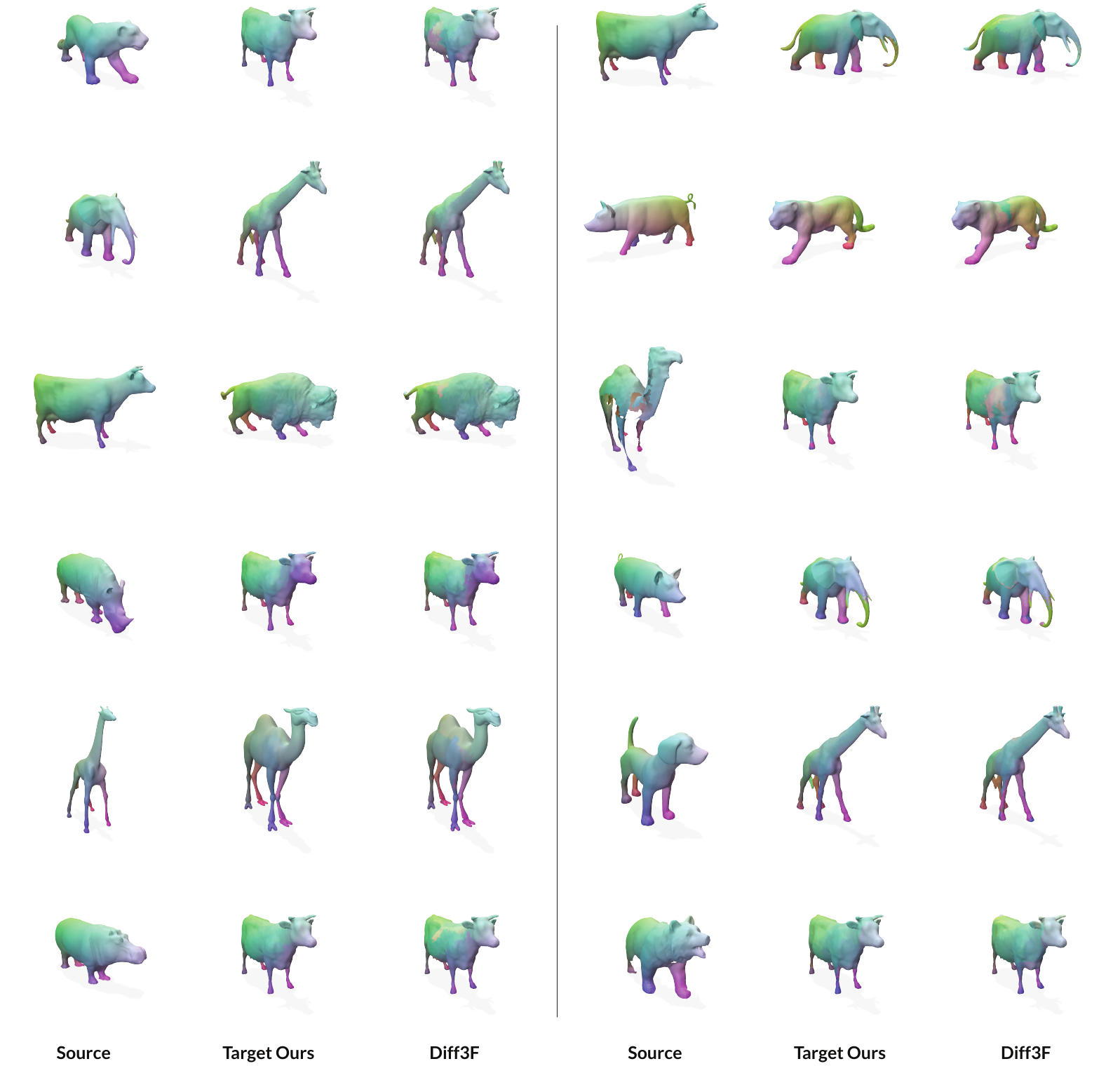}
    \caption{Qualitative results on \dshrecanimals{}. Our features result on average in a more accurate smooth mapping between different limbs.}
    \label{fig:qualitative_shrec20}
\end{figure*}

\refFig{qualitative_shrec20} presents additional results for the \dshrecanimals{} dataset. 
As the dataset only provides $\approx 50$ correspondences for each shape pair, we display the predicted correspondences without dense ground-truth labels.
However, we find that our features generally produce smoother correspondences (e.g., bottom left) and a better separation of legs (e.g., the second to last row on the right).

\subsection{Qualitative Results on ShapeNet}
\label{supp:qualitative_shapenet}
\begin{figure*}
    \centering
    \includegraphics[width=\textwidth]{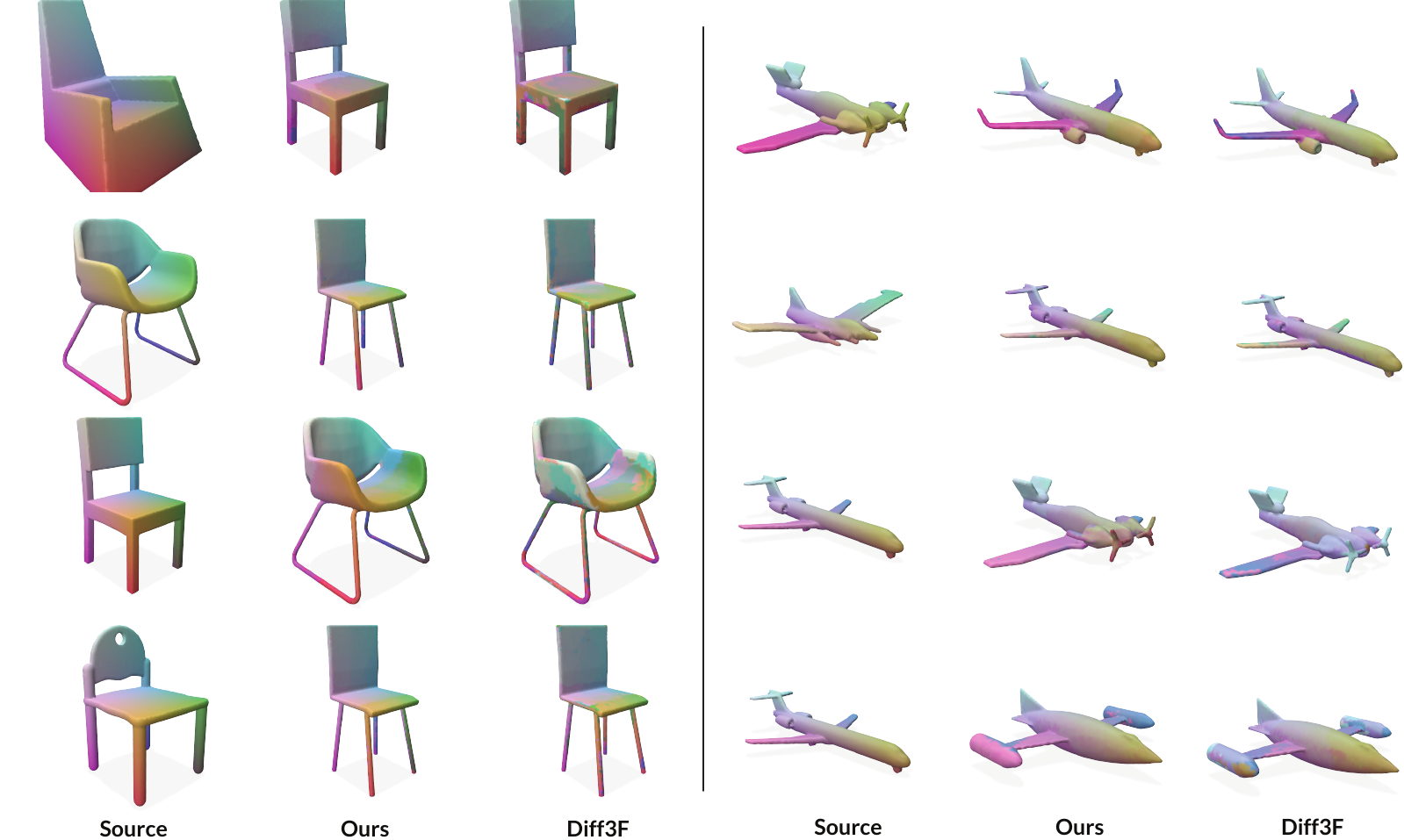}
    \caption{Qualitative results on ShapeNet~\cite{chang2015shapenet}. Our features result on average in a more accurate smooth mapping between chair legs and wings.}
    \label{fig:qualitative_shape}
\end{figure*}

In \refFig{qualitative_shape}, we show 
additional results for chairs and airplanes from ShapeNet~\cite{chang2015shapenet}. 
Since no dense ground truth labels are available, we show the predicted correspondences alone. 
We find that our \emph{surface-aware features} achieve results better than the Diff3F baseline when separating the chair legs (see the top left row) and the airplane wings (see the top right row).

\subsection{Texturing}
\label{supp:texturing}
\begin{figure*}
    \centering
    \includegraphics[width=\textwidth]{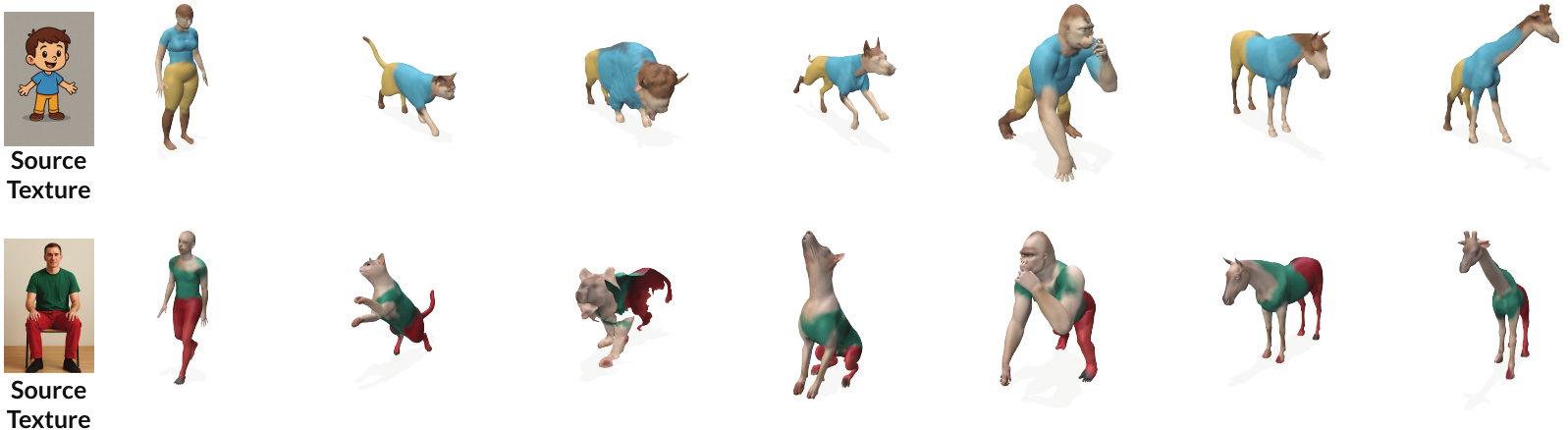}
    \caption{Texturing of 3D meshes from \damass{} and \dtosca{}, based on a 2D image generated with ChatGPT. The appearance is transferred by establishing correspondence between the image features and 3D mesh features. Our method performs well even on incomplete meshes (notice the bear in the second row).}
    \label{fig:2d_texturing_addiitonal}
\end{figure*}

We provide additional examples of 2D-to-3D texturing based on our own features in \refFig{2d_texturing_addiitonal}.

\begin{table}[!h]
\centering
\caption{Correspondence metrics measured for human and animal test shapes. The \textit{Specialized encoder} was trained on a join set of humans and animals following our point-to-point correspondence experiment procedure in \refSec{exp_correspondences}, and thus, the values match \refTbl{results_diff3f} in the paper (see \textit{Ours}). \textit{All-shape encoder} was trained on a larger more generalized training set covering humans and animals but also chairs and airplanes.}

\small
\begin{tabular}{lrrrr}
\toprule
Encoder &   & SHREC’19 & TOSCA & SHREC’20 \\
\midrule
\multirow[c]{2}{*}{Specialized} & err \textdownarrow & \textbf{0.43\footnotesize{±0.76}} & 1.65\footnotesize{±2.15} & \textbf{23.89\footnotesize{±8.90}} \\
 & acc \textuparrow & \textbf{28.78\footnotesize{±9.30}} & \textbf{29.35\footnotesize{±14.53}} & \textbf{73.97\footnotesize{±26.47}} \\
\cline{1-5}
\multirow[c]{2}{*}{All-shape} 
& err \textdownarrow & 0.56\footnotesize{±1.03} & \textbf{1.62\footnotesize{±2.08}} & 4.37\footnotesize{±9.47} \\
& acc \textuparrow & 27.53\footnotesize{±9.47} & 27.45\footnotesize{±14.56} & 70.33\footnotesize{±24.86} \\
\cline{1-5}
\end{tabular}

\label{tbl:one_encoder_for_all}
\end{table}

\subsection{Using a Shared Encoder for All Shapes}
\label{supp:shared_encoder}
In \refSec{experiments}, we train a shared encoder for (\dsurreal{}) and animals (\dsmal{}) shapes as well as separate encoders for chair and airplanes (\textit{ShapeNet}).
Here, we follow the same procedure and train a new single shared encoder on a union of all these shapes and test it on human and animal shapes as in the paper \refTbl{results_diff3f} to assess further generalization of our approach.
The correspondence metrics in \refTbl{one_encoder_for_all} show that the all-shape encoder generally slightly under-performs the specialized encoders but it still improves upon the baselines (see \refTbl{results_diff3f} in the paper).
\dtosca{}{} is an exception, as the larger combined train set marginally reduces the error.

\subsection{Ablating the Number of Training Shapes}

In \refTbl{ablation_training_shape_number}, we explore how the number of training shapes affects the correspondence error with the same fixed validation set and training policy as in the main experiments. 
We vary the number of training samples while retaining a constant animal-to-human shape ratio.
We find that just 2 training samples already decrease the error when compared to Diff3F. 
As expected, additional samples lead to further improvements for SHREC’19 and TOSCA. 
This trend is more subtle for SHREC’20.

\begin{table}[]
\centering
\caption{Training set size (columns) vs. correspondence error.}
\small
\begin{tabular}{llllll}
\hline
         & Diff3f          & 2              & 10             & 50            & 94        \\ \hline
SHREC’19 & 1.69±1.44   & 1.32±1.22    & 1.31±1.31   & 0.48±0.85   & 0.43±0.76 \\ \hline
TOSCA  & 4.51±5.48 & 3.75±3.50 & 2.60±2.74 & 1.84±2.47 & 1.65±2.15 \\ \hline
SHREC’20 & 5.34±10.22 & 3.89±8.49 & 4.05±9.90 & 3.96±9.33 & 3.89±8.90 \\ \hline
\end{tabular}

\label{tbl:ablation_training_shape_number}
\end{table}

\end{document}